\documentclass[10pt,journal,compsoc]{IEEEtran}
\usepackage{times}
\usepackage[nocompress]{cite}
\usepackage{amsmath}
\usepackage{amsfonts,amssymb,times}
\usepackage{acronym}
\usepackage[linesnumbered,ruled,vlined]{algorithm2e}
\usepackage{algpseudocode}
\usepackage{algcompatible}
\usepackage{array}
\usepackage{tabularx}
\usepackage{multirow}
\usepackage{setspace}
\usepackage{mdwmath}
\usepackage{mdwtab}
\usepackage{eqparbox}
\usepackage{graphicx}
\usepackage{subfigure}
\usepackage{psfrag}
\usepackage{url}
\usepackage{flushend}
\usepackage{bm}
\selectfont

\newcommand{\argmin}{\operatornamewithlimits{\mathrm{arg\,min}}}

\begin{document}

\title{Structured Inverted-File k-Means Clustering\\
	for High-Dimensional Sparse Data}
\author{Kazuo~Aoyama and Kazumi~Saito 
\IEEEcompsocitemizethanks{\IEEEcompsocthanksitem K. Aoyama is with 
NTT Communication Science Laboratories, Kyoto, Japan.\protect\\
E-mail: kazuo.aoyama.rd@hco.ntt.co.jp
\IEEEcompsocthanksitem K. Saito is with Kanagawa University.\protect\\
E-mail: k-saito@kanagawa-u.ac.jp
}
}

\IEEEtitleabstractindextext{%
\begin{abstract}
This paper presents an architecture-friendly
k-means clustering algorithm called SIVF
for a large-scale and high-dimensional sparse data set.
Algorithm efficiency on time is often measured by the number of 
costly operations such as similarity calculations.
In practice, however, it depends greatly on 
how the algorithm adapts to an architecture of the computer system 
which it is executed on.
Our proposed SIVF employs 
invariant centroid-pair based filter (ICP) 
to decrease the number of similarity calculations
between a data object and centroids of all the clusters.
To maximize the ICP performance, 
SIVF exploits for a centroid set an inverted-file 
that is structured so as to reduce pipeline hazards.
We demonstrate in our experiments on real large-scale document data sets
that SIVF operates at higher speed and with lower memory consumption 
than existing algorithms.
Our performance analysis reveals that 
SIVF achieves the higher speed by suppressing performance degradation factors of 
the number of cache misses and  branch mispredictions 
rather than less similarity calculations.
\end{abstract}

}
\maketitle

\IEEEpeerreviewmaketitle
\ifCLASSOPTIONcompsoc
\IEEEraisesectionheading{\section{Introduction}\label{sec:intro}}
\else
\section{Introduction}\label{sec:intro}
\fi
Machine learning algorithms are required to efficiently process 
huge data sets in many applications 
with increasing an amount of available data \cite{marsland}.
Such data sets are often high-dimensional sparse ones, 
which are ubiquitous, e.g., 
text (image or audio) data with bag-of-words 
(-visual or -acoustic words) representations \cite{buttcher,yang,sivic,george} 
and log data in computational advertising and 
recommender systems \cite{aggarwal}. 
To design and implement an efficient algorithm for the data sets,
it is important to leverage advantages of a modern computer system 
which the algorithm is executed on.
In general,
algorithm efficiency is measured by computational complexity
on time and space regarding increasing input sizes
so as to be platform-independent and instance-independent \cite{kleinberg}.
However, the relative merits of algorithms on speed performance 
may turn out to be different in practice 
from those evaluated based on the efficiency measure
since their performance depends on characteristics of 
the data sets and the computer systems. 

A modern computer system contains two main components: 
processors and a hierarchical memory system.
A processor has several operating units each of which has 
deep pipelines with superscalar out-of-order execution 
and multilevel cache hierarchy \cite{hennepat}.
The memory system consists of registers and multilevel caches 
in a processor and external memories such as a main memory
and flash storages \cite{zhang}.
To efficiently operate an algorithm at high throughput in such a system, 
we must prevent {\em pipeline hazards}, 
which cause the pipeline stalls resulting in degrading the pipeline performance.
A serious hazard is a control hazard induced by 
branch mispredictions \cite{evers,eyerman}.
Another is a data hazard that can occur
when data dependence exists between instructions. 
In particular, the data hazard caused by cache misses leads to 
serious performance degradation (e.g., \cite{chen}).
To designing an {\em architecture-friendly} algorithm, 
we have to suppress both branch mispredictions and cache misses.

To design an architecture-friendly algorithm 
for large-scale and high-dimensional sparse data sets,
we consider a widely-used Lloyd-type $k$-means clustering algorithm \cite{wu}
because the algorithm is one of fundamental machine learning algorithms 
and has been improved on speed performance by reducing costly similarity calculations
based on the foregoing efficiency measure \cite{hamerly-book}.
Lloyd's algorithm \cite{lloyd,macqueen}, 
which is an iterative heuristic algorithm, 
partitions a given object data set into $k$ subsets (clusters)
with given positive integer $k$. 
By repeating two steps of an assignment and an update step 
until convergence from a given initial state,
the algorithm locally minimizes an objective function, which is 
defined by 
the sum of the squared Euclidean distances between all pairs of 
an object feature vector and a mean feature vector of the cluster 
to which the object is assigned.
Many accelerated Lloyd's algorithms have also been reported
as described in Section~\ref{subsec:lloyd}.

There is a special Lloyd-type algorithm for a text data set, 
a spherical $k$-means algorithm \cite{dhillon}.
Unlike the Lloyd's algorithm,
the spherical $k$-means uses feature vectors normalized by their $L_2$ norms, 
i.e., points on a unit hypersphere, as an input data set  
and adopts a cosine similarity for a similarity measure between a pair of points.
Each mean feature vector is also normalized by its $L_2$ norm.
An objective function is defined by 
the sum of the cosine similarities between all the pairs of 
an object feature vector and a mean feature vector of the cluster 
to which the object is assigned.
A solution by the spherical $k$-means coincides with that 
by the Lloyd's algorithm that uses the same feature vectors 
although their similarity and distance measures differ from each other.
We employ the same settings as that of the spherical $k$-means
to design our proposed algorithm dealing with high-dimensional sparse data sets
like text data sets in Section~\ref{sec:prop}.

Our challenge is to develop 
a high-performance Lloyd-type $k$-means clustering algorithm
for a large-scale and high-dimensional sparse data set, 
exploiting advantages of the architecture in the modern computer system.
We propose a structured inverted-file $k$-means clustering algorithm
referred to as {\em SIVF}.
Our proposed {\em SIVF} utilizes sparse expressions 
for both object and mean feature vectors 
for low memory consumption 
and 
applies an inverted-file data structure to the {\em mean} feature vectors.
For high-speed performance, {\em SIVF} leverages 
an invariant centroid-pair based filter (ICP) that reduces 
similarity calculations. 
The inverted-file in {\em SIVF} has a special structure that 
enables the ICP to work efficiently, resulting in 
reducing branch mispredictions and last-level cache misses 
as shown in Sections~\ref{sec:exp} and \ref{sec:disc}.

Our contributions are threefold:
\begin{enumerate}
\item 
We present a simple yet efficient architecture-friendly 
$k$-means clustering algorithm, 
a structured inverted-file $k$-means clustering algorithm ({\em SIVF}), 
for a large-scale and high-dimensional sparse data set 
with potentially numerous classes in Section \ref{sec:prop}.
Our proposed {\em SIVF} utilizes a structured inverted file
for a set of mean feature vectors
to make an invariant centroid-pair based filter (ICP) work efficiently.
\item
We experimentally demonstrate that {\em SIVF}
achieves superior performance on speed and memory consumption
when it is applied to large-scale and high-dimensional 
real document data sets with large $k$ values,
comparing it with existing algorithms.
\item
We analyze the {\em SIVF} performance with the {\em perf tool} \cite{perf}.
The analysis reveals that
{\em SIVF}'s high speed is clearly attributed to two main factors: 
fewer cache misses and fewer branch mispredictions.
They are detailed in Sections~\ref{sec:exp} and \ref{sec:disc}.
\end{enumerate}

The remainder of this paper consists of the following six sections.
Section \ref{sec:relate} briefly reviews related work 
from viewpoints that clarify the distinct aspects of our work.
Section \ref{sec:pre} describes preliminaries for understanding our proposed algorithm.
Section \ref{sec:prop} explains our proposed {\em SIVF} in detail.
Section \ref{sec:exp} shows our experimental settings and demonstrates the results.
Section \ref{sec:disc} discusses {\em SIVF}'s performance.
The final section provides our conclusion.

\section{Related Work}\label{sec:relate}
Our algorithm is an accelerated Lloyd-type algorithm suitable to
a large-scale sparse data set.
This section reviews acceleration algorithms, followed by
algorithms employing inverted-file structure for sparse data.

\subsection{Acceleration Algorithms}\label{subsec:lloyd}
A $k$-means clustering problem is defined as follows.
Given a set of object feature vectors that are points
in a $D$-dimensional Euclidean space, 
${\cal X}\!=\!\{ \bm{x}_1, \bm{x}_2,\cdots,\bm{x}_N \}$, 
$|{\cal X}|\! =\! N$, $\bm{x}_i\! \in\! \mathbb{R}^D$, 
and a positive integer of $k$,
a $k$-means clustering problem is a problem of finding 
a set of $k$ clusters, 
${\cal C}^*\!=\!\{ C^*_1,C^*_2,\cdots,C^*_k \}$: 
\begin{equation}
{\cal C}^* = \argmin_{{\cal C}=\{ C_1,\cdots, C_k\}}
\left(\: \sum_{C_j\in{\cal C}}\:\sum_{\bm{x}_i\in C_j}
\|\bm{x}_i-\bm{\mu}_j \|_2^2 \:\right)\:, 
\label{eq:obj_funct}
\end{equation}
where $\|\!\star\!\|_2$ denotes the $L_2$ norm of vector $\star$ 
and $\bm{\mu}_j\! \in\! \mathbb{R}^D$ is 
the mean feature vector of cluster $C_j$.
Solving
Eq.~(\ref{eq:obj_funct}) 
is difficult 
in practical use due to a high computational cost. 
Instead, 
Lloyd's algorithm \cite{lloyd,macqueen} finds a local minimum 
in an iterative heuristic manner.
The algorithm repeats two steps of an assignment and an update step
until the convergence or a predetermined termination condition is satisfied.

\begin{algorithm}[!t]
  \caption{~Lloyd-type algorithm at the $r$th iteration}
  \label{algo:lloyd}
	\DontPrintSemicolon
	\KwIn{${\cal X}$,~~
		${\cal M}^{[r-1]}\!=\!\{\bm{\mu}_1^{[r-1]},\cdots,\bm{\mu}_k^{[r-1]}\}$,~~($k$)}
	\KwOut{${\cal C}^{[r]}\!=\!\{ C_1^{[r]},C_2^{[r]},\cdots,C_k^{[r]}\}$,~
				${\cal M}^{[r]}$}
	$C_{j}^{[r]}\leftarrow\emptyset$~,~~$j=1,2,\cdots,k$~\;
	\tcp*[l]{\small Assignment step}
	\ForAll{$\bm{x}_i \in {\cal X}$~}{
		$d_{min}\leftarrow d(\bm{x}_i,\bm{\mu}_{a(\bm{x}_i)}^{[r-1]})\!=\!
			\| \bm{x}_i -\bm{\mu}_{a(\bm{x}_i)}^{[r-1]} \|_2$ \label{algo:lloyd_dmin}\;
		\framebox[45.7mm][c]{$\rm(\,I\,)$} \label{algo:lloyd_1}\;
		\ForAll{$\bm{\mu}_j^{[r-1]} \in {\cal M}^{[r-1]}$}{
			\If{$d(\bm{x}_i,\bm{\mu}_j^{[r-1]})< d_{min}$~}{
				$d_{min}\leftarrow d(\bm{x}_i,\bm{\mu}_j^{[r-1]})$~~and~~
						$a(\bm{x}_i)\leftarrow j$
			}
		}
		$C_{a(\bm{x}_i)}^{[r]}\leftarrow C_{a(\bm{x}_i)}^{[r]}\cup\{ \bm{x}_i\}$\;
	}
	\tcp*[l]{\small Update step}
	$\bm{\mu}_j^{[r]}\leftarrow 
			\left( \sum_{\bm{x}_i\in C_j^{[r]}} \bm{x}_i \right)\,/\,|\,C_j^{[r]}\,|$,
			~~$j=1,2,\cdots,k$\;
	\framebox[50mm][c]{$\rm(I\hspace{-.15em}I)$} \label{algo:lloyd_2}\;
	\Return{${\cal C}^{[r]}\!=\!\{ C_1^{[r]},C_2^{[r]},\cdots,C_k^{[r]}\}$,~
				${\cal M}^{[r]}$}
\end{algorithm}

Algorithm~\ref{algo:lloyd} shows an overview 
of a Lloyd-type algorithm at the $r$th iteration.
The assignment step 
assigns a point represented by object feature vector $\bm{x}_i$ 
to cluster $C_j$ whose centroid 
(mean at the previous iteration $\bm{\mu}_j^{[r-1]}$) 
is closest to $\bm{x}_i$.
At line~\ref{algo:lloyd_dmin}, $d_{min}$ denotes a tentative minimum distance 
from $\bm{x}_i$ to the centroids and 
$a(\bm{x}_i)$ is a function of $\bm{x}_i$ that returns closest
centroid ID $j$.
The update step 
calculates mean feature vector 
$\bm{\mu}_j^{[r]}\!\in\!{\cal M}^{[r]}$ at the $r$th iteration 
using object feature vectors $\bm{x}_{i}\!\in\! C_j^{[r]}$.

Many acceleration algorithms have been reported
\cite{elkan,hamerly,drake,ding,newling,hattori,aoyama}.
We focus on their main filters and review the algorithms 
based on the triangle inequality,
which are compared with our proposed {\em SIVF} in Section~\ref{sec:exp}.
We also describe an invariant centroid-pair based filter (ICP)
\cite{kauko,hattori,bottesch} that {\em SIVF} leverages. 

Main filters of Elkan's \cite{elkan}, Hamerly's \cite{hamerly}, Drake's \cite{drake}, 
and Ding's algorithm \cite{ding} are based on the same principle of skipping 
unnecessary distance calculations.
Elkan's algorithm sets its main filter at line~\ref{algo:lloyd_1} $\rm(\,I\,)$
in Algorithm~\ref{algo:lloyd} as follows.
\begin{eqnarray}
&&\mbox{\bf if}~~
d({\bm x}_i, {\bm \mu}_{a({\bm x}_i)}^{[r-1]}) < 
d_{LB}({\bm x}_i, {\bm \mu}_j^{[r-1]}) -\delta({\bm \mu}_j^{[r-1]}) \nonumber\\
&& \mbox{{\bf then}~~continue}~,\nonumber
\end{eqnarray}
where $d_{LB}({\bm x}_i, {\bm \mu}_j^{[r-1]})$ denotes the lower bound on 
the distance between ${\bm x}_i$ and ${\bm \mu}_j$ at the $(r\!-\!1)$th iteration,
i.e., $d({\bm x}_i, {\bm \mu}_j^{[r-1]})$, and 
$\delta({\bm \mu}_j^{[r-1]})\!=\! d({\bm \mu}_j^{[r-1]}, {\bm \mu}_j^{[r-2]})$.
This algorithm needs the memory capacity of $O(N\!\cdot\!k)$ to store
the distance lower bounds.

Hamerly's algorithm improves Elkan's on memory consumption 
from $O(N\!\cdot \!k)$ to $O(N)$ 
at the expense of a weaker filter.
A main filter is set at line~\ref{algo:lloyd_1}. 
\begin{eqnarray}
&&\mbox{\bf if}~~
d({\bm x}_i, {\bm \mu}_{a({\bm x}_i)}^{[r-1]}) < 
d_{LB}({\bm x}_i, {\bm \mu}_{2nd({\bm x}_i)}^{[r-1]}) -\delta_{max}(\ast) \nonumber\\
&& \mbox{{\bf then}~~continue}~,\nonumber
\end{eqnarray}
where ${\bm \mu}_{2nd({\bm x}_i)}^{[r-1]}$ is the second closest mean to ${\bm x}_i$
at the $(r\!-\! 1)$th iteration and 
$\delta_{max}(\ast)\!=\!\max_{j\neq a_{({\bm x}_i)}} \delta({\bm \mu}_j^{[r-1]})$.

Drake's and Ding's algorithm enhance filtering performance using multiple 
distance lower bounds instead of only one for ${\bm x}_i$ in Hamerly's algorithm.
Drake's algorithm uses $b$ distance lower bounds ($1\!<\! b\!<\! k$) 
between ${\bm x}_i$ and its $b$ closest means. 
The first $(b\!-\! 1)$ lower bounds are determined in the same way as Elkan's algorithm
and the last one is done like Hamerly's.
Ding's algorithm divides $k$ means into $g$ groups ($1\!<\! g \!<\! k$) and 
uses one distance lower bound for each group.
The lower bounds are obtained in the same manner as Hamerly's.
In the limits of ($b,g\rightarrow 1$) and ($b,g\rightarrow k$), 
the corresponding algorithms nearly approach Hamerly's and Elkan's algorithm, 
respectively.
Both the filters are set at line~\ref{algo:lloyd_1}.
The lower bounds in the foregoing algorithms are updated 
at line~\ref{algo:lloyd_2} $\rm{(I\hspace{-.15em}I)}$.
 
We select Drake's and Ding's algorithm 
as the algorithms compared with our {\em SIVF} 
due to their high performance. 
Before the comparison, 
we adapt them to sparse data sets in Section~\ref{subsec:acc}.

ICP 
omits the distance calculations 
between the ${\bm \mu}_j$ and the ${\bm x}_i$ when 
$\delta({\bm \mu}_j^{[r-1]})\!=\!0$ and 
$\delta({\bm \mu}_{a({\bm x}_i)}^{[r-1]})\!=\!0$.
ICP also saves the computational resources because of memory capacity of only $O(k)$ for 
Boolean flags that store whether ${\bm \mu}_j^{[r-1]}$, $j\!=\!1,\cdots,k$, are invariant or not.
We may relax the above restriction on ${\bm x}_i$ as 
$d({\bm x}_i,{\bm \mu}_{a({\bm x}_i)}^{[r-1]})\!\leq\! d({\bm x}_i,{\bm \mu}_{a({\bm x}_i)}^{[r-2]})$
\cite{bottesch} if the memory capacity of $O(k\!+\!N)$ is allowed.

We incorporate ICP to an inverted-file based $k$-means clustering algorithm 
for acceleration.
The inverted-file based algorithm is shown in Section~\ref{subsec:ivf} 
and a na\"{i}ve acceleration algorithm is designed 
for comparison in Section~\ref{subsec:cbicp}.

\subsection{Inverted-File Based Algorithms}\label{subsec:ivf}
\begin{figure}[t]
\begin{center}
	\subfigure[{\normalsize Standard structure}]{
		\psfrag{M}[c][c][0.85]{$N$}
		\psfrag{A}[c][c][0.82]{$\hat{\bm x}_1$}
		\psfrag{B}[c][c][0.82]{$\hat{\bm x}_i$}
		\psfrag{C}[c][c][0.82]{$\hat{\bm x}_N$}
		\includegraphics[width=40mm]{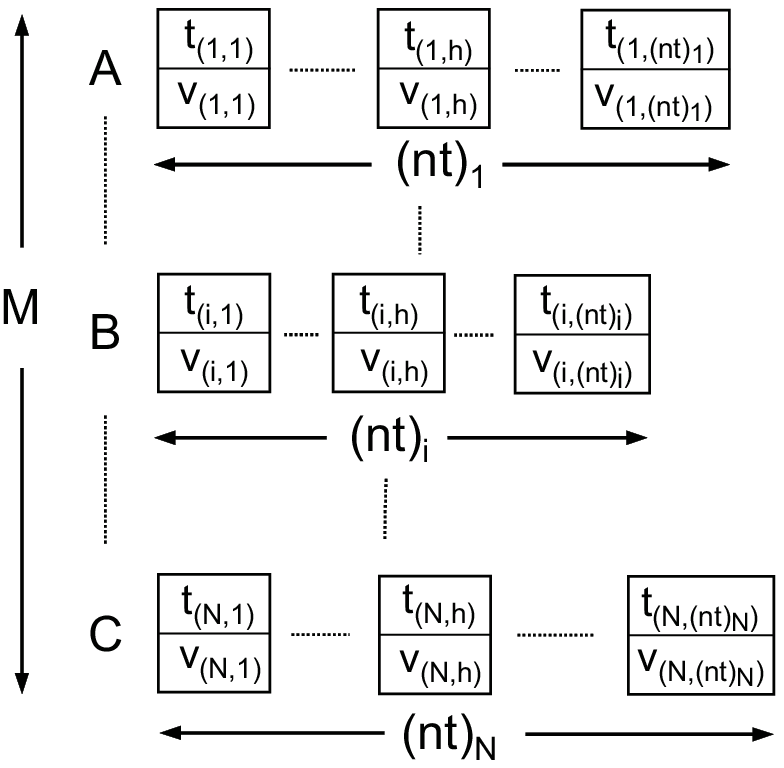}}
	\hspace{3mm}
	\subfigure[{\normalsize Inverted-file data structure}]{
		\psfrag{W}[c][c][0.85]{$D$}
		\psfrag{P}[c][c][0.82]{$\breve{\bm y}_1$}
		\psfrag{Q}[c][c][0.82]{$\breve{\bm y}_s$}
		\psfrag{R}[c][c][0.82]{$\breve{\bm y}_D$}
		\includegraphics[width=40mm]{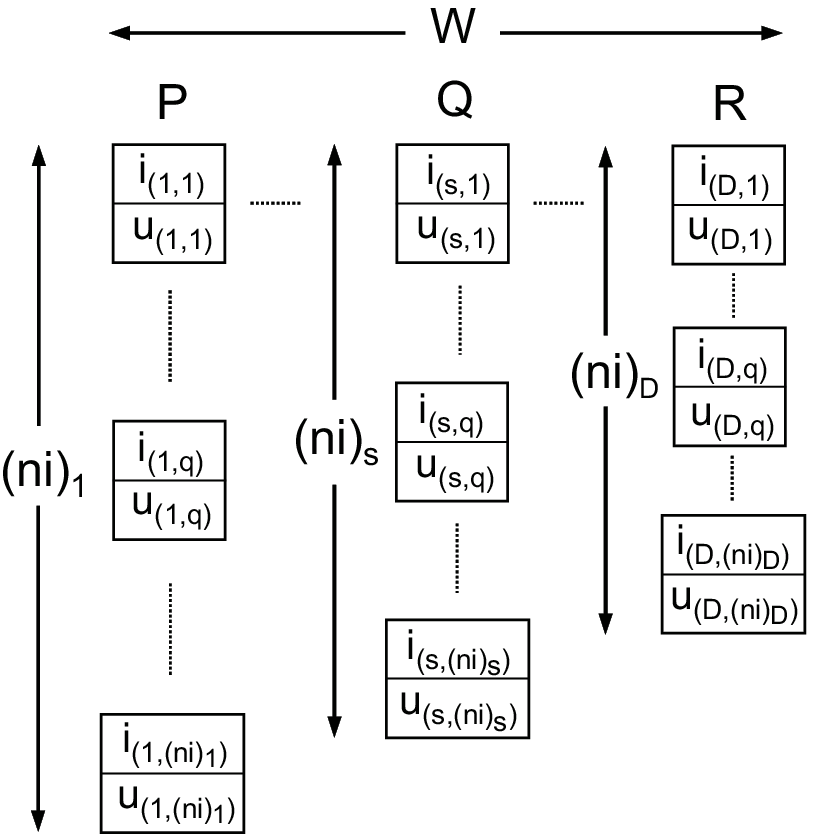} }
\vspace*{-3mm}
\end{center}
\caption{Sparse expressions of feature vectors:
(a) Object feature vector 
$\hat{\bm x}_i\!=\! (t_{(i,h)},v_{(i,h)})$ in standard structure
where $i$ denotes the global ID of objects ($i\!=\!1,2,\cdots,N$) and
$h$ the local index of features ($h\!=\! 1,2,\cdots,(nt)_i$); 
(b) Object array $\breve{\bm y}_s$ that consists of tuples of
object ID $i_{(s,q)}$ and feature $u_{(s,q)}$ in inverted-file data structure
where $s$ denotes the global ID of features ($s\!=\!1,2,\cdots,D$) and
$q$ the local index of objects ($q\!=\! 1,2,\cdots,(ni)_s$).
}
\label{fig:str}
\end{figure}
When designing an algorithm for a sparse data set 
where each object is represented as a sparse feature vector,
we have to carefully determine both a structure of the data set and 
an expression of the feature vector.
Suppose that two object sets are given, 
each of which contains feature vectors normalized by their $L_2$ norms.
The feature vector is a point on the unit hypersphere and 
a similarity between the feature vectors is measured by cosine similarity, i.e.,
their inner-product.
When we calculate a similarity between feature vectors 
each of which has only several non-zero elements,
we need to select a pair of a data structure and an expression of the object sets.
The data structure is either a standard or an inverted-file one 
and 
the expression is either a full or a sparse one.
A sparse expression in a standard data structure is defined 
as a sequence of tuples of a feature ID ($t_{(i,h)}$) and a feature value 
($v_{(i,h)}$) for each object in Fig.~\ref{fig:str}(a).
By contrast, in an inverted-file data structure, 
an object array is defined as a sequence of tuples of an object ID 
($i_{(s,q)}$) and the corresponding feature value ($u_{(s,q)}$) 
for each feature in Fig.~\ref{fig:str}(b).

In search algorithms for a text data set, 
a pair of a sparse expression and an inverted-file data structure 
is often adopted
for invariant database that contains a set of object feature vectors 
\cite{samet,harman,knuth,zobel,buttcher} as in Fig.~\ref{fig:str}(b).
Given query feature vector $\{\hat{\bm x}_1\}$ represented with 
a sparse expression in a standard data structure in Fig.~\ref{fig:str}(a),
a search algorithm identifies the $(nt)_1$ object arrays 
by the feature IDs of $t_{(1,h)}$, $h\!=\!1,\cdots,(nt)_1$, 
and calculates only the products of $v_{(1,h)}$ and $u_{(s,q)}$ 
where $s\!=\!t_{(1,h)}$ for their similarity.
Thus it can find preferable documents quickly from the inverted-file database.

A $k$-means clustering algorithm processes both a data object set and 
a mean (centroid) set.
If the data object set is a large-scale and high-dimensional sparse one,
it is natural that 
the object feature vector is represented with a sparse expression.
Furthermore, it is advisable that the mean feature vector is also represented 
with the sparse expression.
In this situation, there are two usages of the inverted-file structure.
One is to apply the inverted-file structure to an object data set \cite{broder}.
The other is to do it to a mean set, 
which is referred to as {\em IVF}.
Since the latter leads to higher performance than the former
when applied to large-scale and high-dimensional real document data sets \cite{ivf}, 
we adopt the latter one for our proposed {\em SIVF}
as in Section~\ref{sec:prop}.

\section{Preliminaries}\label{sec:pre}
This section describes both a way for applying the acceleration algorithms 
in Section~\ref{subsec:lloyd} 
to sparse data sets and a na\"{i}ve method for accelerating inverted-file 
based $k$-means clustering algorithm by ICP. 
Our proposed algorithm is compared with the acceleration algorithms adapted 
with the former way and the algorithm made with the latter method.
\subsection{Applying Accelerations to Sparse Data}\label{subsec:acc}
Drake's and Ding's algorithm were originally designed 
for low- to moderate-dimensional dense data sets \cite{drake, ding} 
below 1,000 dimensions
such as MNIST handwritten digit dataset (784 dimensions) \cite{mnist} 
and 80 million tiny images (384 dimensions) \cite{tiny}.
We adapt them to a high-dimensional sparse data set whose 
dimensionality is over 100,000 and which has several non-zero elements.
We assume that the standard data structure is applied to both 
an object data set and a mean set.
It is natural that the object feature vectors are represented with 
the sparse expression 
because of the object data size and its sparsity as in Fig.~\ref{fig:str}(a).
Then there are two choices to represent the mean feature vectors: 
the sparse and the full expression

Let us suppose that the mean feature vectors are represented with 
the sparse expression.
This representation provides a positive effect on memory consumption
while it causes speed-performance degradation. 
When both the object and the mean feature vectors employ the sparse expression
in the standard data structure, i.e., 
each feature element in the vectors is a tuple of a feature ID and a feature value,
an algorithm detects a pair of the tuples with an identical feature ID 
in the object and the mean feature vectors using many conditional branches.
Since it is difficult to predict truth values of the conditions in most cases,
a lot of branch mispredictions occur, resulting in the performance degradation.

In the case of the full expression,
the foregoing conditional branches are unnecessary.
A feature vector represented with the full expression is a sequence of 
$D$ feature values arranged in ascending order of feature ID from 1 to $D$.
If a feature value at a feature ID does not exist, 
zero-padding is performed at the ID.
For this expression,
an algorithm can directly access a feature value in the mean feature vector 
by using the object feature ID.
We adopt the full expression for the mean feature vectors 
to adapt Drake's and Ding's algorithm to high-dimensional sparse data sets
although this expression has the drawbacks of 
the requirement of a large amount of memory capacity and
the possibility of decreasing an effective cache-hit rate.

\subsection{Accelerating Inverted-File k-Means}\label{subsec:cbicp}
\begin{algorithm}[!t]
  \caption{Assignment step in {\em IVF} at $r$th iteration} 
  \label{algo:ivf_assign}
	\DontPrintSemicolon
	\KwIn{$\hat{\cal X}$,~~$\breve{\cal M}^{[r-1]}$,~~($k$)}
	\KwOut{${\cal C}^{[r]}\!=\!\{ C_1^{[r]},C_2^{[r]},\cdots,C_k^{[r]}\}$}
	\BlankLine
	$C_{j}^{[r]}\leftarrow\emptyset$~,~~$j=1,2,\cdots,k$~\;
	\ForEach{$\hat{\bm{x}}_i\!=\!(t_{(i,h)},v_{(i,h)})_{h=1}^{(nt)_i}\in \hat{\cal X}$~~
		\label{algo:ivf_sim0}}{
		$\rho_{max}\!\leftarrow\! 0$,~~
		$\bm{\rho}\!=\!(\rho_1,\rho_2,\cdots,\rho_j,\cdots,\rho_k)\!\leftarrow\!\bm{0}$\;
		$S_i\!=\!\{t_{(i,1)},~t_{(i,2)},~\cdots,~t_{(i,h)},~\cdots,~t_{(i,(nt)_i)}\}$\;
		\ForAll{$s\!\leftarrow\!t_{(i,h)}\in S_i$}{
			\tcp*[l]{\small 
				$\breve{\bm{\xi}}_s^{[r-1]}\!=\![(c_{(s,q)},u_{(s,q)})_{q=1}^{(mf)_s}]^{[r-1]}$
			}
			\ForAll{$(c_{(s,q)},u_{(s,q)})^{[r-1]}\in \breve{\bm \xi}_s^{[r-1]}$}{
				$\rho_{c_{(s,q)}}\leftarrow \rho_{c_{(s,q)}}+v_{(i,h)}\!\times\! u_{(s,q)}$
				\label{algo:ivf_sim1}
			}
		}
		\For{$j\!\leftarrow\! 1$ \KwTo $k$}{
			\lIf{ $\rho_j\!>\! \rho_{max}$ }{
				$\rho_{max}\!\leftarrow\!\rho_j$~and~$a(\hat{\bm{x}}_i)\!\leftarrow\! j$}
		}
		$C_{a(\hat{\bm{x}}_i)}^{[r]}\leftarrow 
			C_{a(\hat{\bm{x}}_i)}^{[r]}\cup\{ \hat{\bm{x}}_i\}$\;
	}
	\Return{${\cal C}^{[r]}$}
\end{algorithm}

Algorithm~\ref{algo:ivf_assign} shows a pseudocode of the assignment step 
at $r$th iteration in {\em IVF} \cite{ivf}.
Similarities are calculated in the triple loop 
at lines~\ref{algo:ivf_sim0} to \ref{algo:ivf_sim1}, where
$\breve{\bm \xi}_s^{r-1]}$ and $(mf)_s$ are the $s$th mean (centroid) array 
and the maximum of the $s$th local index that correspond to 
$\breve{y}_s$ and $(ni)_s$ in Fig.~\ref{fig:str}(b), respectively. 
{\em IVF} calculates similarities between object $\hat{\bm x}_i$ 
and only the limited centroids listed in centroid arrays $\breve{\bm \xi}_s^{[r-1]}$.
On the negative side, the listed centroids are always the targets of 
similarity calculations without any filters.

To reduce the similarity calculations like other acceleration algorithms, 
we try to incorporate the invariant centroid-pair based filter (ICP) 
to {\em IVF}.
Other acceleration algorithms based on the triangle inequality
directly use the relationship between an object and a centroid while
ICP exploits the relationship between a pair of centroids 
as shown in Section~\ref{subsec:lloyd}.
That is, ICP does not calculate the similarity between 
${\bm \mu}_j^{[r-1]}$ of $\delta({\bm \mu}_j^{[r-1]})\!=\!0$ and 
$\hat{\bm x}_i$ of $\delta({\bm \mu}_{a(\hat{\bm x}_i)}^{[r-1]})\!=\!0$.
We prepare a Boolean flag $\lambda_j$, $j\!=\!1,\cdots,k$ for each centroid, 
which is 1 if $\delta({\bm \mu}_j^{[r-1]})\!=\!0$, otherwise 0.
A na\"{i}ve way to design {\em IVF} with ICP is to use two conditional branches 
based on the Boolean flag.

\begin{algorithm}[!t]
  \caption{{\em IVF-CBICP}: Part of assignment step}
  \label{algo:cbicp_assign}
	\DontPrintSemicolon
	\ForEach{$\hat{\bm{x}}_i\!=\!(t_{(i,h)},v_{(i,h)})_{h=1}^{(nt)_i}\in \hat{\cal X}$~~}{
			$\rho_{max}\!\leftarrow\! 0$,~~
			$\bm{\rho}\!=\!(\rho_1,\rho_2,\cdots,\rho_j,\cdots,\rho_k)\!\leftarrow\!\bm{0}$\;
			$S_i\!=\!\{t_{(i,1)},~t_{(i,2)},~\cdots,~t_{(i,h)},~\cdots,~t_{(i,(nt)_i)}\}$\;
			\eIf{~\framebox[18mm][l]{~$\lambda_{a(\hat{\bm{x}}_i)}^{[r-1]} =\!1$}~
				\label{algo:cbicp_cb}}{
				\ForAll{$s\!\leftarrow\!t_{(i,h)}\in S_i$}{
					\tcp*[l]{\small 
					$\breve{\bm{\xi}}_s^{[r-1]}\!=\![(c_{(s,q)},u_{(s,q)})_{q=1}^{(mf)_s}]^{[r-1]}$}
					\For{$q\leftarrow 1$ \KwTo \framebox[13.8mm][l]{~$(mf)_s$}\label{algo:cbicp_innerloop0}}{
						\If{\framebox[18mm][l]{~$\lambda_{c_{(s,q)}}^{[r-1]} =\!0$}~
							\label{algo:cbicp_deepcb}}{
							$\rho_{c_{(s,q)}}\leftarrow \rho_{c_{(s,q)}}+v_{(i,h)}\!\times\! u_{(s,q)}$\;
							\label{algo:cbicp_innerloop1}
						}
					}
				}
			}{
				\ForAll{$s\!\leftarrow\!t_{(i,h)}\in S_i$}{
				\For{$q\!\leftarrow\! 1$ \KwTo \framebox[13.8mm][l]{~$(mf)_s$}}
				{$\rho_{c_{(s,q)}}\leftarrow \rho_{c_{(s,q)}}+v_{(i,h)}\!\times\! u_{(s,q)}$\;}
			}
		}
	}
\end{algorithm}

Algorithm~\ref{algo:cbicp_assign} shows a part of the assignment step in 
{\em IVF} with ICP using conditional branches referred to as {\em IVF-CBICP},
which corresponds to lines~\ref{algo:ivf_sim0} to \ref{algo:ivf_sim1} in 
Algorithm~\ref{algo:ivf_assign}.
Just before the second inner-most loop at line~\ref{algo:cbicp_cb}, 
the Boolean flag $\lambda_{a(\hat{\bm x}_i)}^{[r-1]}$ is evaluated.
If the flag's value is 1, i.e., the cluster which $\hat{\bm x}_i$ belongs to is invariant,
only the moving centroids are targets for the similarity calculations.
The centroids are selected with the conditional branch in the inner-most loop 
at line~\ref{algo:cbicp_deepcb}.
Thus {\em IVF-CBICP} utilizes ICP with two conditional branches.
Although {\em IVF-CBICP} successfully reduced similarity calculations,
it got little improvement on the speed performance as detailed in 
Section~\ref{sec:disc}.

\section{Proposed Algorithm: SIVF}\label{sec:prop}
We propose a structured inverted-file $k$-means clustering algorithm 
({\em SIVF}\,).
This algorithm was designed to efficiently process 
a large-scale and high-dimensional sparse data set at high speed 
and with low memory consumption, in particular, when a large $k$ value is given,
i.e., under a tough condition where any algorithms need a lot of computational resources.
First, we adopted an inverted-file data structure for a centroid (mean) set \cite{ivf}
to leverage a physical memory effectively.
Second, we incorporated an invariant centroid-pair based filter (ICP) \cite{kauko,hattori,bottesch}
to reduce a computational cost by skipping unnecessary similarity calculations.
Last, to combine ICP with a tentative algorithm using the inverted file,
we gave the inverted file a special structure, 
which enables {\em SIVF} to exploit ICP without the conditional branch 
at line~\ref{algo:cbicp_deepcb} in Algorithm~\ref{algo:cbicp_assign}.
The structured inverted file consists of two parts:
the front and the back part contain moving and invariant centroids, respectively.
By replacing the end point of $(mf)_s$ 
at line~\ref{algo:cbicp_deepcb} in Algorithm~\ref{algo:cbicp_assign}
with an index at the boundary of the foregoing two parts,
we can remove the conditional branch.
The pseudocode is shown in 
Algorithms~\ref{algo:sivf_assign} and \ref{algo:sivf_update}.

\subsection{Assignment Step}\label{subsec:assign}

\begin{algorithm}[!t]
  \caption{{\em SIVF} assignment step at the $r$th iteration} 
  \label{algo:sivf_assign}
	\DontPrintSemicolon
	\SetKwBlock{DoParallel}{do in parallel}{end}
	\KwIn{$\hat{\cal X}$,~~$\breve{\cal M}^{*[r-1]}$,~~
			$\bm{\lambda}^{[r-1]}\!=\!(\lambda_1^{[r-1]},\cdots,\lambda_k^{[r-1]})$,~~($k$)}
	\KwOut{${\cal C}^{[r]}\!=\!\{ C_1^{[r]},C_2^{[r]},\cdots,C_k^{[r]}\}$,~~${\bm\lambda^{[r]}}$}
	\BlankLine
	$C_{j}^{[r]}\leftarrow\emptyset$~,~~$j=1,2,\cdots,k$~\;
	\DoParallel{
		\tcp*[l]{\small Calculate similarities}
		\ForEach{$\hat{\bm{x}}_i\!=\!(t_{(i,h)},v_{(i,h)})_{h=1}^{(nt)_i}\in \hat{\cal X}$~~}{
				$\rho_{max}\!\leftarrow\! 0$,~~
				$\bm{\rho}\!=\!(\rho_1,\rho_2,\cdots,\rho_j,\cdots,\rho_k)\!\leftarrow\!\bm{0}$\;
				$S_i\!=\!\{t_{(i,1)},~t_{(i,2)},~\cdots,~t_{(i,h)},~\cdots,~t_{(i,(nt)_i)}\}$\;
				\eIf{~\framebox[19mm][l]{~$\lambda_{a(\hat{\bm{x}}_i)}^{[r-1]}\!=\!1$}~}{
					\ForAll{$s\!\leftarrow\!t_{(i,h)}\in S_i$}{
						\tcp*[l]{\small 
							$\breve{\bm{\xi}}_s^{*[r-1]}\!=\![(c_{(s,q)},u_{(s,q)})_{q=1}^{(mf)_s}]^{*[r-1]}$}
						\For{$q\leftarrow 1$ \KwTo \framebox[15.8mm][l]{~$(mf_{[0]})_s$}}
						{ \tcp*[l]{\small $(mf_{[0]})_s$ inclusive}
						$\rho_{c_{(s,q)}}\leftarrow \rho_{c_{(s,q)}}+v_{(i,h)}\!\times\! u_{(s,q)}$\;
						}
					}
				}{
					\ForAll{$s\!\leftarrow\!t_{(i,h)}\in S_i$}{
						\For{$q\!\leftarrow\! 1$ \KwTo \framebox[13.8mm][l]{~$(mf)_s$}}
						{$\rho_{c_{(s,q)}}\leftarrow \rho_{c_{(s,q)}}+v_{(i,h)}\!\times\! u_{(s,q)}$\;}
					}
				}
			\tcp*[l]{\small Assign $\hat{\bm{x}}_i$ to a cluster}
			\For{$j\!\leftarrow\! 1$ \KwTo $k$}{
				\lIf{ $\rho_j\!>\! \rho_{max}$ }{
					$\rho_{max}\!\leftarrow\!\rho_j$~and~$a(\hat{\bm{x}}_i)\!\leftarrow\! j$}
			}
			$C_{a(\hat{\bm{x}}_i)}^{[r]}\leftarrow 
				C_{a(\hat{\bm{x}}_i)}^{[r]}\cup\{ \hat{\bm{x}}_i\}$\;
		}
	}
	\tcp*[l]{\small Mark invariant clusters}
	$\bm{\lambda}^{[r]}\leftarrow \bm{0}$\;
	\ForAll{ $C_j^{[r]}\in {\cal C}^{[r]}$ }{ 
		\lIf{ $C_j^{[r]} = C_j^{[r-1]}$ }{$\lambda_j^{[r]}\leftarrow 1$}
	}
	\Return{${\cal C}^{[r]}$,~
			$\bm{\lambda}^{[r]}\!=\!(\lambda_1^{[r]},\cdots,\lambda_k^{[r]})$}
\end{algorithm}

Algorithm~\ref{algo:sivf_assign} shows the assignment step in {\em SIVF} 
at the $r$th iteration.
From the results at the $(r\!-\!1)$th iteration, {\em SIVF} receives 
a centroid set and a Boolean-flag vector.
The centroid set, which is the mean set at the $(r\!-\!1)$th iteration, 
is represented with structured inverted-file sparse expression $\breve{\cal M}^{*[r-1]}$.
The Boolean-flag vector ${\bm \lambda}^{[r-1]}$ consists of $k$ elements 
$\lambda_j^{[r-1]}$, $j\!=\!1,2,\cdots,k$.
$\lambda_j^{[r-1]}\!=\!1$ if the members in the $j$th cluster are invariant 
between the $(r\!-\!2)$th and the $(r\!-\!1)$th iteration,
otherwise $\lambda_j^{[r-1]}\!=\!0$.
{\em SIVF} also uses a data object set represented with standard sparse expression 
$\hat{\cal X}$.
At the assignment step, 
each cluster $C_j^{[r]}$ and Boolean-flag vector ${\bm \lambda}_j^{[r]}$ are generated.

The triple loop at lines~3 to 13 and the assignment of an object to a cluster 
at lines~14 to 16 are executed by multithread processing.
At the outer-most loop, the $i$th object feature vector 
($\hat{\bm{x}}_i\!\in\!\hat{\cal X}$) is chosen 
to determine a cluster which the $i$th object belongs to.
$\hat{\bm x}_i$ consists of 
$(nt)_i$ tuples $(t_{(i,h)},v_{(i,h)})$, $h\!=\!1,2,\cdots,(nt)_i$, where 
$(nt)_i$ denotes the number of distinct terms that the $i$th object uses,
$h$ is the local counter, $t_{(i,h)}$ is the global feature ID (term ID) from 1 to $D$,
and $v_{(i,h)}$ is the corresponding feature value such as {\em tf-idf}.

We insert the conditional branch just before the inner double loop 
at line~6 to identify whether the cluster $C_{a(\hat{\bm x}_i)}^{[r-1]}$ 
which the $i$th object belongs to is invariant or not.
If the cluster is invariant, we calculate similarities of the $i$th object to
only the moving centroids that change their positions due to the changes of
the cluster members, otherwise we have to do the similarities to all the centroids.
This is ICP function of skipping the similarity calculations.
To exploit the foregoing ICP, we give inverted-file centroid array 
$\breve{\bm \xi}_s^{[r-1]}$
a simple but effective structure, where 
the moving centroids are placed at the front part indexed by 1 to $(mf_{[0]})_s$ 
in $\breve{\bm \xi}_s^{[r-1]}$.
This centroid array $\breve{\bm \xi}_s^{*[r-1]}$ 
consists of $(mf)_s$ tuples $(c_{(s,q)},u_{(s,q)})^{*[r-1]}$, 
$q\!=\!1,2,\cdots,(mf)_s$, 
where $c_{(s,q)}$ denotes the global centroid ID from 1 to $k$, 
$u_{(s,q)}$ is the corresponding value, and 
$(mf)_s$ denotes the centroid (mean) frequency of term ID $s$.
Note that the centroid array is partitioned into two parts of the front
($1\!\leq\! q\!\leq \!(mf_{[0]})_s$) and the back part 
($(mf_{[0]})_s\!<\! q\!\leq \!(mf)_s$).
Owing to the structured inverted-file centroid array,
we can realize the ICP function only to specify the end position
of the inner-most loop without the conditional branch. 
A partial similarity (corresponding to a partial inner product) 
of the $i$th object to the $c_{(s,q)}$th centroid is calculated 
and stored at $\rho_{c_{(s,q)}}$ at lines~9 and 13.

Just after the inner double loop has been completed,
the $i$th object is assigned to the $a(\hat{\bm{x}}_i)$th cluster 
whose centroid most closely resembles
at lines~14 to 16.
For the next iteration, we mark invariant clusters at lines~17 to 19.
Last, the assignment step passes the cluster set ${\cal C}^{[r]}$ and 
the Boolean-flag vector ${\bm \lambda}^{[r]}$ 
to the following update step.

\subsection{Update Step}\label{subsec:update}

\begin{algorithm}[!t]
	\caption{{\em SIVF} update step at the $r$th iteration} 
	\label{algo:sivf_update}
	\DontPrintSemicolon
	\KwIn{$\hat{\cal X}$,~${\cal C}^{[r]}$,~$\bm{\lambda}^{[r]}$}
	\KwOut{$\breve{\cal M}^{*[r]}\!=\!(\breve{\bm \xi}_1^{*[r]},\breve{\bm \xi}_2^{*[r]},
		\cdots,\breve{\bm \xi}_p^{*[r]},\cdots,\breve{\bm \xi}_D^{*[r]})$}
	\BlankLine
	\tcp*[l]{\small Determine an inverted-file structure}
	$\bm{mf}_{[0]}\!=\!((mf_{[0]})_1,\cdots,(mf_{[0]})_D)\leftarrow \bm{0}$\;
	$\bm{mf}_{[1]}\!=\!((mf_{[1]})_1,\cdots,(mf_{[1]})_D)\leftarrow \bm{0}$\;
	\For{$j\!\leftarrow\! 1$ \KwTo $k$}{
		\eIf{$\lambda_j^{[r]}\!=\!0$}{
			\ForAll{$C_j^{[r]} \in {\cal C}^{[r]}$}{
				$S_{\mu}\leftarrow\emptyset$~~\tcp*[h]{Tentative term ID set}\;
				\lForAll{$\hat{\bm x}_i \in C_j^{[r]}$}{
					$S_{\mu}\leftarrow S_{\mu}\cup \{t_{(i,h)}\!\in\! S_i\}$}
				\lForAll{$s\in S_{\mu}$}{
					$(mf_{[0]})_s\leftarrow (mf_{[0]})_s \!+\! 1$}
			}
		}{
			\ForAll{$C_j^{[r]} \in {\cal C}^{[r]}$}{
				$S_{\mu}\leftarrow\emptyset$~~\tcp*[h]{Tentative term ID set}\;
				\lForAll{$\hat{\bm x}_i \in C_j^{[r]}$}{
					$S_{\mu}\leftarrow S_{\mu}\cup \{t_{(i,h)}\!\in\! S_i\}$}
				\lForAll{$s\in S_{\mu}$}{
					$(mf_{[1]})_s\leftarrow (mf_{[1]})_s \!+\! 1$}
			}
		}
	}
	\lFor{$p\leftarrow 1$ \KwTo $D$}{~$(mf)_p \leftarrow (mf_{[0]})_p +\! (mf_{[1]})_p$}
	\BlankLine
	\tcp*[l]{\small Make structured inverted-file $\breve{\cal M}^{*[r]}$}
	$q_{[0]p}\!\leftarrow\!1$,~
	\framebox[32mm][l]{~$q_{[1]p}\!\leftarrow\!(mf_{[0]})_p\!+\!1$~}~,~
	$p=1,2,\cdots,D$\;
	\ForAll{$C_j^{[r]} \in {\cal C}^{[r]}$}{
		\tcp*[l]{\small Calculate mean features}
		${\bm w}\!=\!(w_1,\cdots,w_D)\!\leftarrow\! {\bm 0}$~\tcp*[h]{Tentative vector}\;
		\ForAll{$\hat{\bm x}_i \in C_j^{[r]}$}{
			\lForAll{$s\!\leftarrow\!t_{(i,h)}\in S_i$}{
				$w_{s}\!\leftarrow\! w_{s}\! +\!v_{(i,h)}$}
		}
		\lFor{$p\leftarrow 1$ \KwTo $D$}{~~$w_p\!\leftarrow\! w_p/|C_j^{[r]}|$}
		\tcp*[l]{\small Make inverted-file mean arrays}
		\eIf{$\lambda_j^{[r-1]}\!=\!0$}{
			\For{$p\leftarrow 1$ \KwTo $D$}{
				\If{$w_p\!\neq\!0$}{
					$c_{(p,q_{[0]p})}\!\leftarrow\! j$,~~$u_{(p,q_{[0]p})}\!\leftarrow\! w_p/\|\bm{w}\|_2$\;
					$q_{[0]p}\!\leftarrow\! q_{[0]p}\!+\!1$\;}
			}
		}{
			\For{$p\leftarrow 1$ \KwTo $D$}{
				\If{$w_p\!\neq\!0$}{
					$c_{(p,q_{[1]p})}\!\leftarrow\! j$,~~$u_{(p,q_{[1]p})}\!\leftarrow\! w_p/\|\bm{w}\|_2$\;
					$q_{[1]p}\!\leftarrow\! q_{[1]p}\!+\!1$\;}
			}
		}
	}
	\Return{$\breve{\cal M}^{*[r]}$}
\end{algorithm}

Algorithm~\ref{algo:sivf_update} shows the update step at the $r$th iteration.
At the update step, we calculate each mean of $k$ clusters based on the object assignment
and make a structured inverted file 
$\breve{\cal M}^{*[r]}$ 
that consists of $D$ inverted-file mean arrays
$\breve{\bm \xi}_p^{*[r]}$, $p\!=\!1,2,\cdots,D$.

We first determine both the length $(mf)_p$ and the end position $(mf_{[0]})_s$
of the inverted-file mean array 
$\breve{\bm \xi}_p^{*[r]}$, $p\!=\!1,2,\cdots,D$ 
at lines~1 to 14.
According to the element $\lambda_j^{[r]}$ in the Boolean-flag vector,
we separately enumerate the numbers of moving and invariant means that 
contains the $s$th term and store those in $(mf_{[0]})_s$ and $(mf_{[1]})_s$,
respectively. 

We initialize two local counters,
$q_{[0]p}$ for moving means and $q_{[1]p}$ for invariant means, 
in the $p$th inverted-file mean array $\breve{\bm \xi}_p^{*[r]}$, 
where $p$ denotes the global term ID.
Next, we calculate a mean feature vector in each cluster $C_j^{[r]}$ 
at lines~17 to 20.
Based on an evaluation result of $\lambda_j^{[r]}$ in the Boolean-flag vector,
we place both cluster ID (mean ID) and its feature value to 
an appropriate position in the inverted-file mean array 
at lines~21 to 30.
Thus we complete structured inverted-file $\breve{\cal M}^{*[r]}$ 
that consists of $\breve{\bm \xi}_p^{*[r]}$, $p\!=\!1,\cdots,D$.

{\em SIVF} utilizes a structured inverted file for a centroid set, 
which fuses ICP that skips unnecessary similarity calculations 
with an ordinary inverted file suitable to processing a large-scale sparse data set.
Therefore, we can expect that {\em SIVF} efficiently works for
a large-scale and high-dimensional sparse data set with low memory consumption
and at high speed.
In the following section, 
we qualitatively evaluate the {\em SIVF} performance, 
comparing it with existing algorithms.

\section{Experiments}\label{sec:exp}
We first describe data sets used in our experiments,
a platform including a computer system 
where the algorithms were executed,
and performance measures for evaluation.
Next, we compare our proposed {\em SIVF} with existing algorithms
regarding performance and analyze their performances with 
the {\em perf tool} \cite{perf}.
We experimentally demonstrate that {\em SIVF} is superior to 
the existing algorithms 
when applied to high-dimensional sparse data sets.

\subsection{Data Sets}\label{subsec:data}
We employed two different types of large-scale and high-dimensional 
sparse real document data sets: 
{\em PubMed Abstracts} (PubMed for short) \cite{pubmed}
and {\em The New York Times Articles} (NYT).

The PubMed data set contains 8,200,000 documents (texts) 
each of which was represented by the term (distinct word) counts.
We made a feature vector normalized by its $L_2$ norm from each document,
which consisted of the {\em tf-idf} values of the corresponding terms.
Each feature vector was regarded as a point on a unit hypersphere.
We made five data sets that were referred to as 500K, 1M, 2M, 5M, and 8M-sized PubMed.
The $N$-sized PubMed had $N$ feature vectors chosen at random without duplication 
from all of the vectors, e.g., 1M-sized PubMed had 1,000,000 feature vectors.
The data sets contained distinct terms (vocabulary) corresponding to dimensionality
of 139,845, 140,914, 141,041, 141,043, and 141,043 in ascending order of data size. 
The average term frequencies in the documents, i.e., 
the average numbers of non-zero elements in the feature vectors, were 58.96, 58.95, 
58.97, 58.96, and 58,96 in the same order.

We extracted 1,285,944 articles from NYT 
from 1994 to 2006 and counted the frequency of the term occurrences 
after stemming and stop word removal.
In the same manner as PubMed, we made a set of feature vectors
with 495,714 dimensionality.
The average number of non-zero elements in the feature vectors was
225.76.
Thus both the data sets are large-scale and high-dimensional sparse ones.

\subsection{Platform and Measures}\label{subsec:platform}
All the algorithms were executed on a computer system 
that was equipped with two Xeon E5-2697v3 2.6-GHz CPUs 
with three-level caches from levels 1 to level 3 (last level) \cite{hammarlund} 
and a 256-GB main memory, 
by multithreading with OpenMP \cite{openmp} within the memory capacity.
In the CPU, the out-of-order superscalar execution was performed 
with eight issue widths
and the last-level cache has 36,700,160 (35M) bytes consisting of 64-byte blocks
with 20-way set associative placement and least-recently used (LRU) replacement 
\cite{hammarlund,jongerius}.
The algorithms were implemented in C and compiled with a GNU C compiler (gcc) 
version 8.2.0 on the optimization level of 
{\sf -O3}. 
The performances of the algorithms were evaluated with 
CPU time (or clock cycles) until convergence and 
the maximum size of the physical memory occupied through the iterations.
To analyze the speed performance, we measured {\em performance degradation factors}
with the {\em perf tool} (Linux profiling with performance counters) \cite{perf}.
In particular, we focused on the number of completed instructions (Inst for short), 
branch mispredictions (BM) , and last-level cache misses (LLCM).

\subsection{Performance Evaluation}\label{subsec:eval}
First, we compared {\em SIVF} with two existing algorithms of 
Drake's ({\em Drake}$^+$) \cite{drake} and 
Ding's ({\em Ding}$^+$) \cite{ding} algorithm 
and our designed {\em Lloyd-ICP}, which was a modified Lloyd's algorithm
incorporating ICP as a baseline.
The three compared algorithms were implemented with the method 
shown in Section~\ref{subsec:acc}.
That is, each mean feature vector was represented 
as a vector with full dimensionality.
Next, we analyzed {\em SIVF} speed performance, focusing on 
the number of similarity calculations and the performance degradation factors
(DFs).
The evaluation results showed that 
{\em SIVF} was superior to the compared algorithms in PubMed and NYT.
The high speed came from the suppression of DFs 
rather than less similarity calculations.
Our results in PubMed is shown here 
and those in NYT is done in Appendix~\ref{sec:appA}.

\subsubsection{Comparison with existing algorithms}\label{subsubsec:exist}

\begin{figure}[t]
\begin{center}\hspace*{1mm}
	\subfigure[{\normalsize Avg. elapsed time}]{
		\psfrag{K}[c][c][0.88]{
			\begin{picture}(0,0)
				\put(0,0){\makebox(0,-5)[c]{Number of clusters: $k$ ($\times\! 10^3$)}}
			\end{picture}
		}
		\psfrag{T}[c][c][0.88]{
			\begin{picture}(0,0)
				\put(0,0){\makebox(0,20)[c]{Avg. elapsed time (sec)}}
			\end{picture}
		}
		\psfrag{Q}[c][c][0.8]{$1$}
		\psfrag{R}[c][c][0.8]{$10$}
		\psfrag{G}[c][c][0.8]{$2$}
		\psfrag{H}[c][c][0.8]{$5$}
		\psfrag{J}[c][c][0.8]{$20$}
		\psfrag{X}[r][r][0.8]{$1$}
		\psfrag{Y}[r][r][0.8]{$10$}
		\psfrag{Z}[r][r][0.8]{$10^2$}
		\psfrag{U}[r][r][0.8]{$10^3$}
		\psfrag{A}[r][r][0.63]{\em SIVF}
		\psfrag{B}[r][r][0.63]{{\em Ding}$^+$}
		\psfrag{C}[r][r][0.63]{{\em Drake}$^+$}
		\psfrag{D}[r][r][0.63]{\em Lloyd-ICP}
		\includegraphics[width=40.2mm]{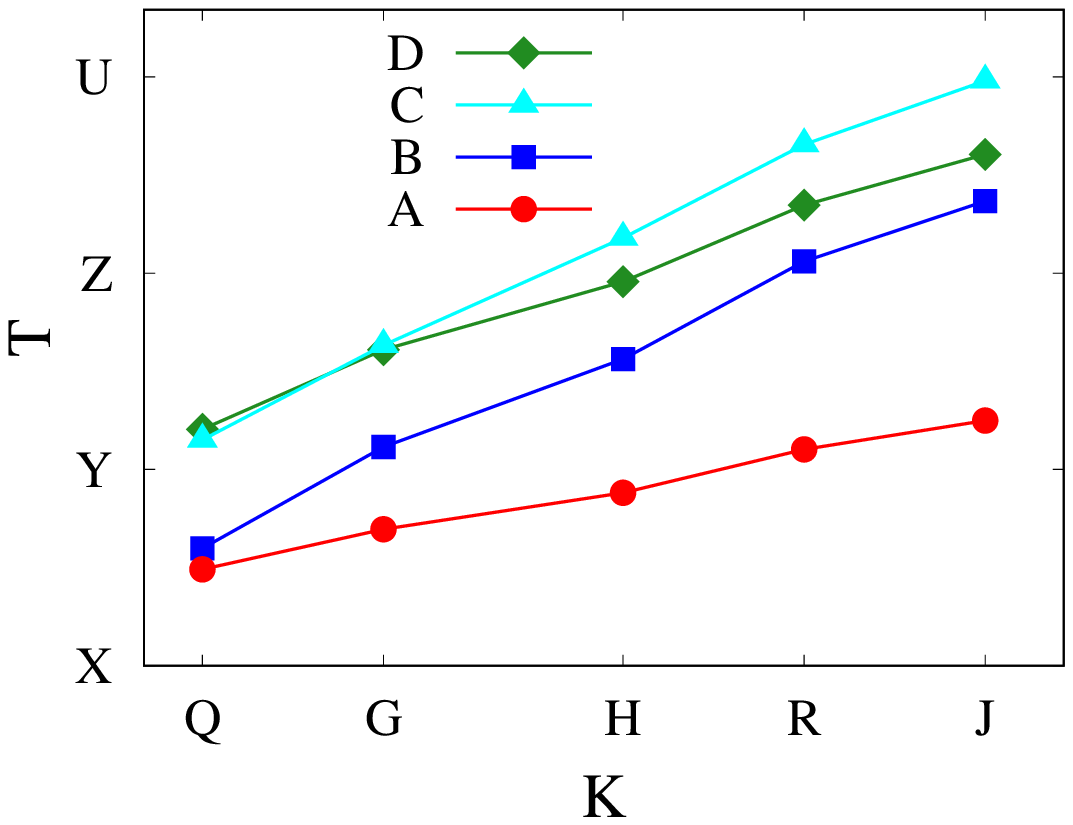}
	} \hspace*{1mm} 
	\subfigure[{\normalsize Max. memory size}]{
		\psfrag{K}[c][c][0.88]{
			\begin{picture}(0,0)
				\put(0,0){\makebox(0,-5)[c]{Number of clusters: $k$ ($\times\! 10^3$)}}
			\end{picture}
		}
		\psfrag{T}[c][c][0.88]{
			\begin{picture}(0,0)
				\put(0,0){\makebox(0,20)[c]{Max. memory size (GB)}}
			\end{picture}
		}
		\psfrag{Q}[c][c][0.8]{$1$}
		\psfrag{R}[c][c][0.8]{$10$}
		\psfrag{G}[c][c][0.8]{$2$}
		\psfrag{H}[c][c][0.8]{$5$}
		\psfrag{J}[c][c][0.8]{$20$}
		\psfrag{X}[r][r][0.8]{$1$}
		\psfrag{Y}[r][r][0.8]{$10$}
		\psfrag{Z}[r][r][0.8]{$10^2$}
		\psfrag{A}[r][r][0.63]{\em SIVF}
		\psfrag{B}[r][r][0.63]{{\em Ding}$^+$}
		\psfrag{C}[r][r][0.63]{{\em Drake}$^+$}
		\psfrag{D}[r][r][0.63]{\em Lloyd-ICP}
		\psfrag{M}[r][r][0.63]{(Data set)}
		\includegraphics[width=40.2mm]{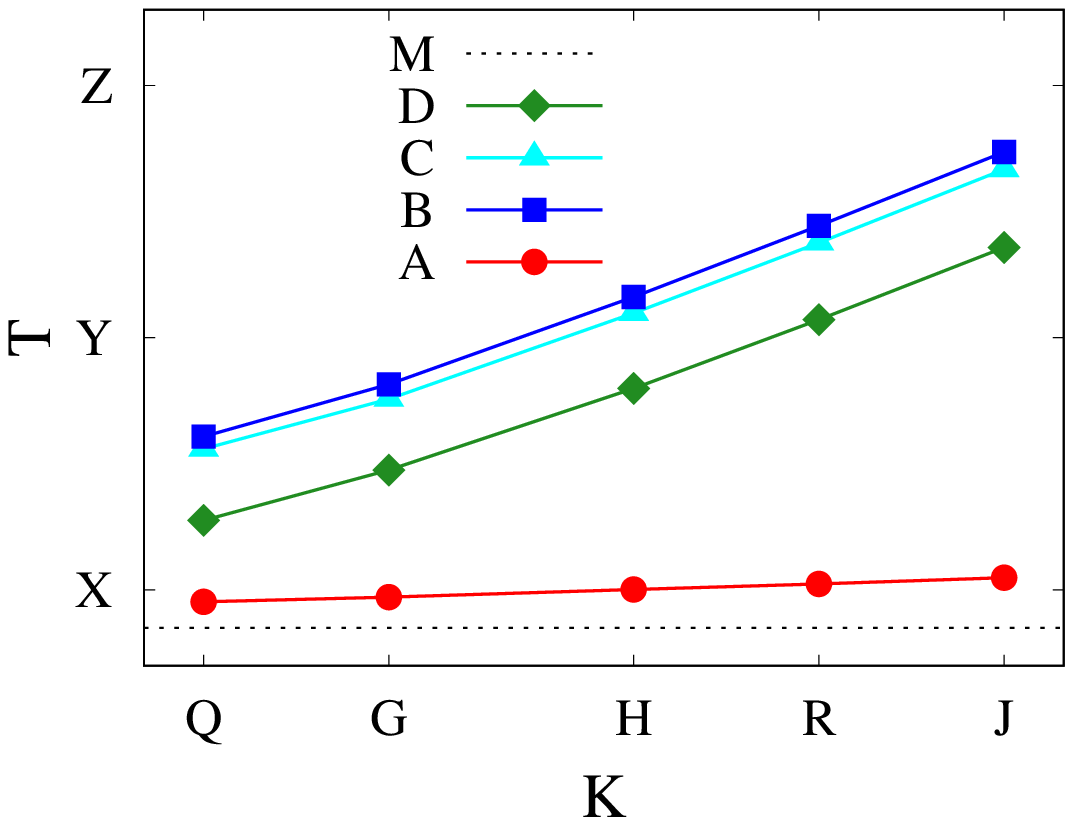}
	}
\end{center}
\vspace*{-3mm}
\caption{Performance of four algorithms executed by 50-thread processing 
with given k in 1M-sized PubMed.
(a) Average elapsed time per iteration and 
(b) Occupied maximum physical memory size through iterations are plotted along k 
with log-log scale. 
}
\label{fig:performK_pubmed}
\vspace*{-2mm}
\end{figure}

Figures~\ref{fig:performK_pubmed}(a) and (b) show that
each of the four algorithms required 
average elapsed time per iteration and maximum physical memory size 
through iterations until convergence when they were executed by 50-thread
processing with OpenMP in 1M-sized PubMed, given the $k$ values of
(1,000, 2,000, 5,000, 10,000, 20,000).
As shown in Section~\ref{subsec:lloyd}, 
the compared Ding's and Drake's algorithm have parameters $g$ and $b$, respectively.
These parameters were set at $k$/10 as shown in \cite{ding}.
The average elapsed time of {\em SIVF} slowly increased with $k$ and 
was much smaller than the others' in the large $k$ region of $k\!\geq\!2,000$,
in particular, it was only 7.6\% of that required by {\em Ding}$^+$ 
at $k\!=\!20,000$.
This $k$ region is a tough condition 
where the algorithms except {\em SIVF} needed so much computational time.
Regarding memory consumption, 
{\em SIVF} used small memory sizes in all the $k$ values
because of its sparse feature vector representation 
of both the data object and mean sets.

\begin{figure}[t]
\begin{center}\hspace*{1mm}
	\begin{tabular}{cc}	
	\subfigure{
		\psfrag{K}[c][c][0.88]{
			\begin{picture}(0,0)
				\put(0,0){\makebox(0,-5)[c]{Data size ($\times 10^6$)}}
			\end{picture}
		}
		\psfrag{T}[c][c][0.88]{
			\begin{picture}(0,0)
				\put(0,0){\makebox(0,20)[c]{Avg. elapsed time (sec)}}
			\end{picture}
		}
		\psfrag{P}[c][c][0.8]{$0.5$}
		\psfrag{Q}[c][c][0.8]{$1$}
		\psfrag{R}[c][c][0.8]{$2$}
		\psfrag{S}[c][c][0.8]{$5$}
		\psfrag{W}[c][c][0.8]{$10$}
		\psfrag{X}[r][r][0.8]{$1$}
		\psfrag{Y}[r][r][0.8]{$10$}
		\psfrag{Z}[r][r][0.8]{$10^2$}
		\psfrag{U}[r][r][0.8]{$10^3$}
		\psfrag{A}[r][r][0.63]{\em SIVF}
		\psfrag{B}[r][r][0.63]{{\em Ding}$^+$}
		\psfrag{C}[r][r][0.63]{{\em Drake}$^+$}
		\psfrag{D}[r][r][0.63]{\em Lloyd-ICP}
		\includegraphics[width=41.5mm]{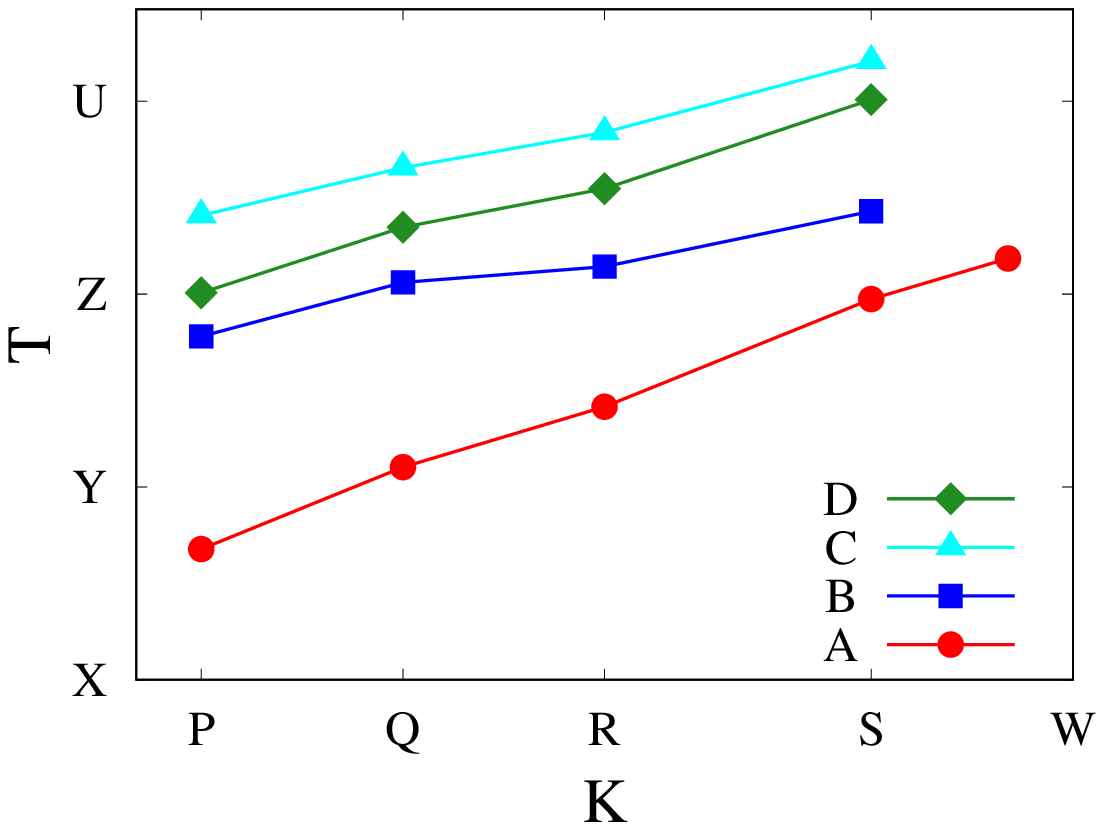}
	} &
	\subfigure{
		\psfrag{K}[c][c][0.88]{
			\begin{picture}(0,0)
				\put(0,0){\makebox(0,-5)[c]{Data size ($\times 10^6$)}}
			\end{picture}
		}
		\psfrag{T}[c][c][0.88]{
			\begin{picture}(0,0)
				\put(0,0){\makebox(0,20)[c]{Max. memory size (GB)}}
			\end{picture}
		}
		\psfrag{P}[c][c][0.8]{$0.5$}
		\psfrag{Q}[c][c][0.8]{$1$}
		\psfrag{R}[c][c][0.8]{$2$}
		\psfrag{S}[c][c][0.8]{$5$}
		\psfrag{W}[c][c][0.8]{$10$}
		\psfrag{X}[r][r][0.8]{$0.1$}
		\psfrag{Y}[r][r][0.8]{$1$}
		\psfrag{Z}[r][r][0.8]{$10$}
		\psfrag{U}[r][r][0.8]{$10^2$}
		\psfrag{A}[r][r][0.63]{\em SIVF}
		\psfrag{B}[r][r][0.63]{{\em Ding}$^+$}
		\psfrag{C}[r][r][0.63]{{\em Drake}$^+$}
		\psfrag{D}[r][r][0.63]{\em Lloyd-ICP}
		\psfrag{E}[r][r][0.63]{\em (Data set)}
		\includegraphics[width=41.5mm]{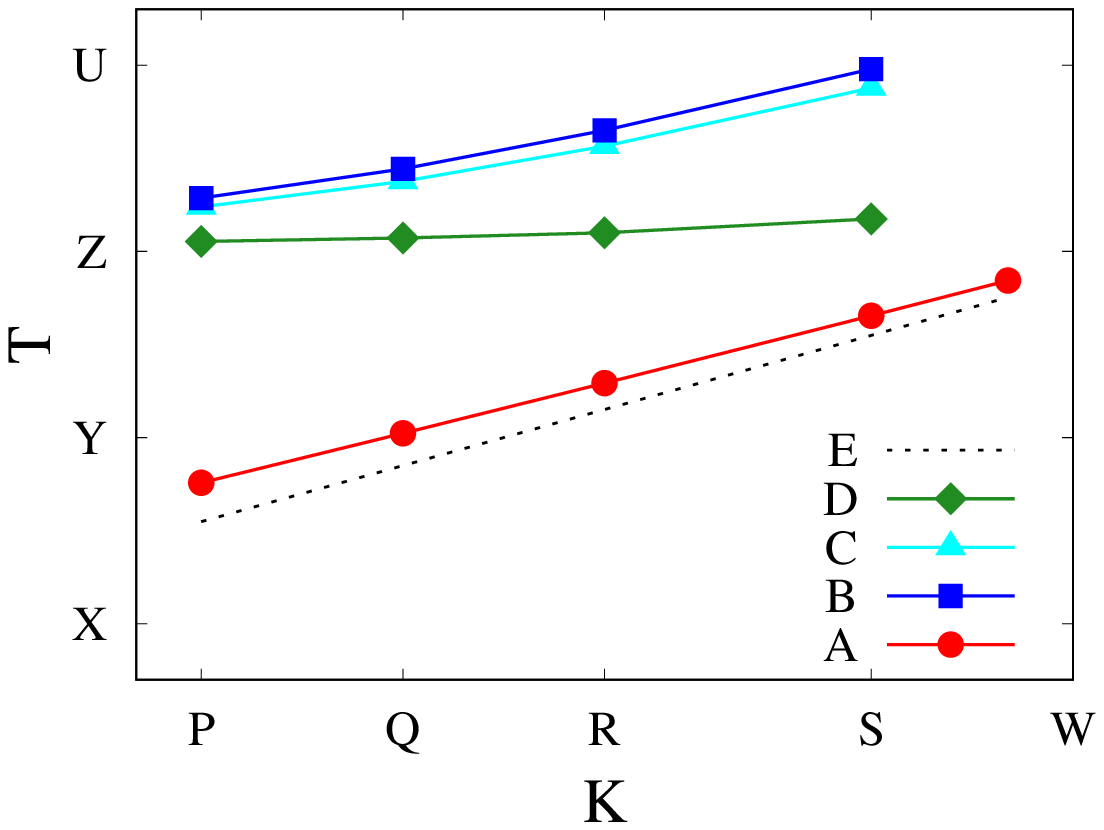}
	}\tabularnewline
	(a)~Avg. elapsed time & (b)~Max. memory size \tabularnewline
	\end{tabular}
\end{center}
\vspace*{-3mm}
\caption{Performance of four algorithms executed by 50-thread processing 
for PubMed with different sizes when k=10,000.
(a) Average elapsed time per iteration and 
(b) Occupied maximum memory size through iterations are plotted along data size
with log-log scale. 
}
\label{fig:performN_pubmed}
\vspace*{-2mm}
\end{figure}

Figures~\ref{fig:performN_pubmed}(a) and (b) show 
the average elapsed time per iteration and the maximum physical memory size 
that each of the four algorithms required with data size $N$ 
when the algorithms were executed in $N$-sized PubMed at $k\!=\!10,000$
by 50-thread processing, 
where $N = (5\!\times\!10^5, 1\!\times\!10^6, 
2\!\times\!10^6, 5\!\times\!10^6, 8\!\times\!10^6)$.
{\em SIVF} achieved the best performance among the algorithms.
The algorithms except {\em SIVF} did not work for 8M-sized PubMed.

\subsubsection{Performance analysis}\label{subsubsec:analy}

\begin{figure}[t]
\begin{center}
	\psfrag{K}[c][c][0.95]{Iteration}
	\psfrag{T}[c][c][0.95]{
		\begin{picture}(0,0)
			\put(0,0){\makebox(0,22)[c]{Elapsed time (sec)}}
		\end{picture}
	}
	\psfrag{P}[c][c][0.86]{$0$}
	\psfrag{Q}[c][c][0.86]{$10$}
	\psfrag{R}[c][c][0.86]{$20$}
	\psfrag{S}[c][c][0.86]{$30$}
	\psfrag{X}[r][r][0.86]{$10$}
	\psfrag{Y}[r][r][0.86]{$10^2$}
	\psfrag{Z}[r][r][0.86]{$10^3$}
	\psfrag{U}[r][r][0.86]{$10^4$}
	\psfrag{V}[r][r][0.86]{$10^5$}
	\psfrag{A}[r][r][0.7]{\em SIVF}
	\psfrag{B}[r][r][0.7]{{\em Ding}$^+$}
	\psfrag{C}[r][r][0.7]{{\em Drake}$^+$}
	\psfrag{D}[r][r][0.7]{\em Lloyd-ICP}
	\includegraphics[width=53mm]{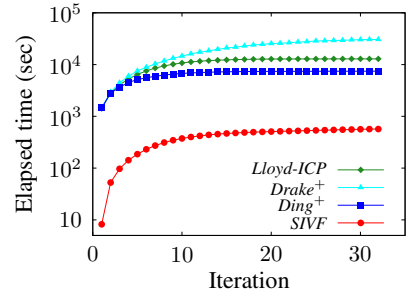}
\end{center}
\vspace*{-3mm}
\caption{
Elapsed time that four algorithms with k=20,000 required until convergence 
when they were executed by 50-thread processing for 1M-sized PubMed.
Elapsed time is plotted along iteration with linear-log scale.
}
\label{fig:itrTime_pubmed}
\vspace*{-2mm}
\end{figure}

\begin{figure}[t]
\begin{center}
	\psfrag{K}[c][c][0.95]{Iteration}
	\psfrag{W}[c][c][0.95]{
		\begin{picture}(0,0)
			\put(0,0){\makebox(0,30)[c]{Normalized \# sim. calc.}}
		\end{picture}
	}
	\psfrag{P}[c][c][0.86]{$0$}
	\psfrag{Q}[c][c][0.86]{$10$}
	\psfrag{R}[c][c][0.86]{$20$}
	\psfrag{S}[c][c][0.86]{$30$}
	\psfrag{X}[r][r][0.86]{$10^{-5}$}
	\psfrag{Y}[r][r][0.86]{$10^{-4}$}
	\psfrag{Z}[r][r][0.86]{$10^{-3}$}
	\psfrag{H}[r][r][0.86]{$10^{-2}$}
	\psfrag{J}[r][r][0.86]{$10^{-1}$}
	\psfrag{L}[r][r][0.86]{$1$}
	\psfrag{A}[r][r][0.7]{\em SIVF}
	\psfrag{B}[r][r][0.7]{{\em Ding}$^+$}
	\psfrag{C}[r][r][0.7]{{\em Drake}$^+$}
	\psfrag{D}[r][r][0.7]{\em Lloyd-ICP}
	\includegraphics[width=53mm]{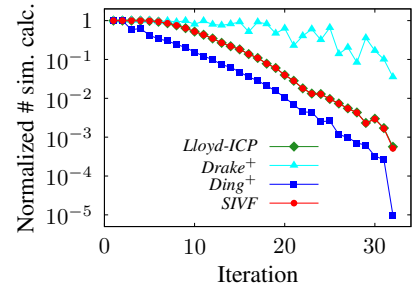}
\end{center}
\vspace*{-3mm}
\caption{
Number of similarity calculations normalized by (N x k) per iteration, 
where N indicates the data size of 1M, 
when four algorithms with k=20,000 were executed by 50-thread processing 
for 1M-sized PubMed.
Normalized number is plotted along iteration with linear-log scale.
}
\label{fig:cal_pubmed}
\vspace*{-2mm}
\end{figure}

To analyze the speed performance of the algorithms, 
we focused on the results in 1M-sized PubMed, given $k\!=\!20,000$,
where the marked performance differences were found.

Figure~\ref{fig:itrTime_pubmed} shows that
the elapsed times that the four algorithms required at each iteration
from the start to the convergence (through 32 iterations)
when they were executed by 50-thread processing.
We notice that {\em SIVF} operated much faster than the others.
The elapsed time of {\em SIVF} was only 568 sec 
while that of the second fastest {\em Ding}$^+$ was 7440 sec.

Figure~\ref{fig:cal_pubmed} shows the filter performance of each algorithm, i.e., 
the ability of skipping unnecessary similarity calculations at each iteration.
The filter performance was evaluated by a rate of the number of similarity calculations 
to $(N\!\times\!k)$, which corresponds to that required by Lloyd's algorithm, and
is illustrated along iteration with linear-log scale.
When a filter works better, its rate becomes smaller.
We notice that the filter of {\em Ding}$^+$ reduced more 
similarity calculations, which is an indicator of a computational cost, 
than the others'.
Why did Ding's algorithm equipped with the high-performance filter 
need more elapsed time 
than {\em SIVF} as shown in Fig.~\ref{fig:itrTime_pubmed}?

\begin{figure}[t]
\begin{center}
	\hspace*{1mm}
	\subfigure[{\normalsize Similarity calculation}]{
		\psfrag{K}[c][c][0.88]{
			\begin{picture}(0,0)
				\put(0,0){\makebox(0,-5)[c]{Number of clusters: $k$ ($\times\! 10^3$)}}
			\end{picture}
		}
		\psfrag{W}[c][c][0.88]{
			\begin{picture}(0,0)
				\put(0,0){\makebox(0,22)[c]{Normalized \# sim. cal.}}
			\end{picture}
		}
		\psfrag{Q}[c][c][0.8]{$1$}
		\psfrag{R}[c][c][0.8]{$10$}
		\psfrag{G}[c][c][0.8]{$2$}
		\psfrag{H}[c][c][0.8]{$5$}
		\psfrag{J}[c][c][0.8]{$20$}
		\psfrag{X}[r][r][0.8]{$0$}
		\psfrag{Y}[r][r][0.8]{$0.2$}
		\psfrag{Z}[r][r][0.8]{$0.4$}
		\psfrag{U}[r][r][0.8]{$0.6$}
		\psfrag{S}[r][r][0.8]{$0.8$}
		\psfrag{T}[r][r][0.8]{$1.0$}
		\psfrag{A}[r][r][0.63]{\em SIVF}
		\psfrag{B}[r][r][0.63]{{\em Ding}$^+$}
		\psfrag{C}[r][r][0.63]{{\em Drake}$^+$}
		\psfrag{D}[r][r][0.63]{\em Lloyd-ICP}
		\includegraphics[width=40mm]{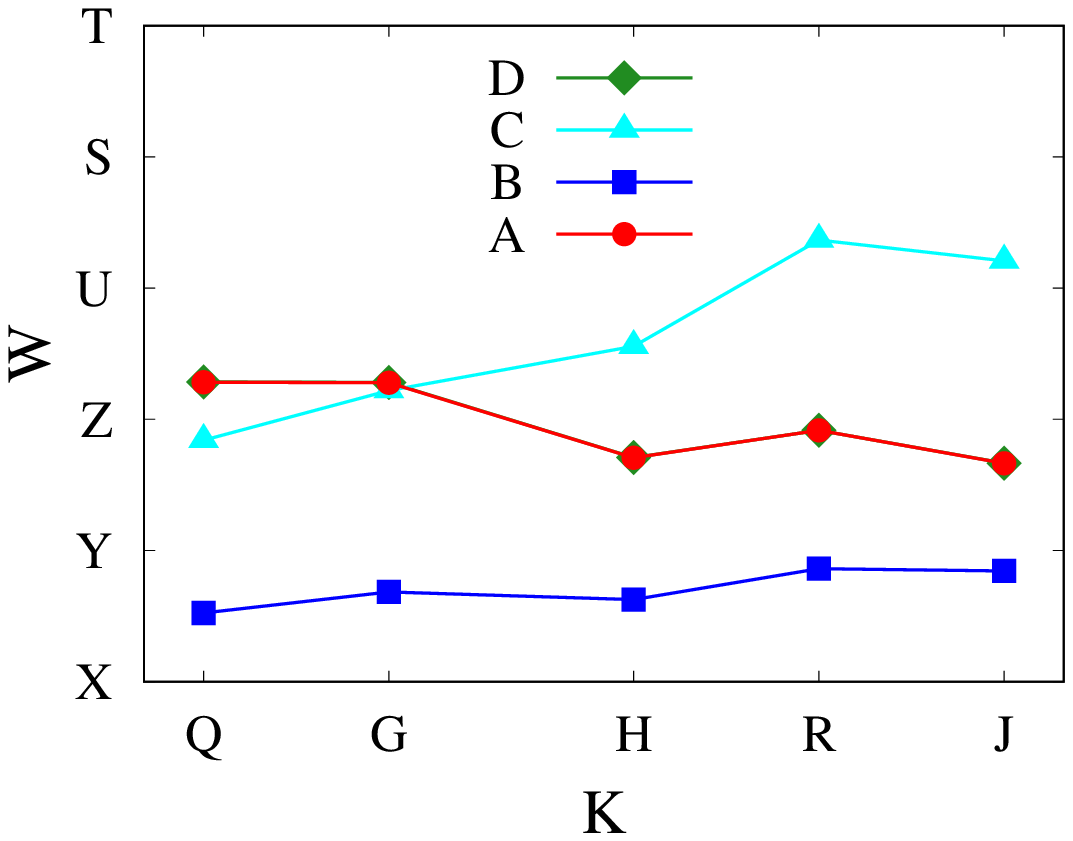}
	}\hspace*{1mm}
	\subfigure[{\normalsize Instruction}]{
		\psfrag{K}[c][c][0.88]{
			\begin{picture}(0,0)
				\put(0,0){\makebox(0,-5)[c]{Number of clusters: $k$ ($\times\! 10^3$)}}
			\end{picture}
		}
		\psfrag{W}[c][c][0.88]{
			\begin{picture}(0,0)
				\put(0,0){\makebox(0,32)[c]{\# instructions}}
			\end{picture}
		}
		\psfrag{Q}[c][c][0.8]{$1$}
		\psfrag{R}[c][c][0.8]{$10$}
		\psfrag{G}[c][c][0.8]{$2$}
		\psfrag{H}[c][c][0.8]{$5$}
		\psfrag{J}[c][c][0.8]{$20$}
		\psfrag{Y}[r][r][0.8]{$10^{11}$}
		\psfrag{Z}[r][r][0.8]{$10^{12}$}
		\psfrag{U}[r][r][0.8]{$10^{13}$}
		\psfrag{S}[r][r][0.8]{$10^{14}$}
		\psfrag{A}[r][r][0.63]{\em SIVF}
		\psfrag{B}[r][r][0.63]{{\em Ding}$^+$}
		\psfrag{C}[r][r][0.63]{{\em Drake}$^+$}
		\psfrag{D}[r][r][0.63]{\em Lloyd-ICP}
		\includegraphics[width=40mm]{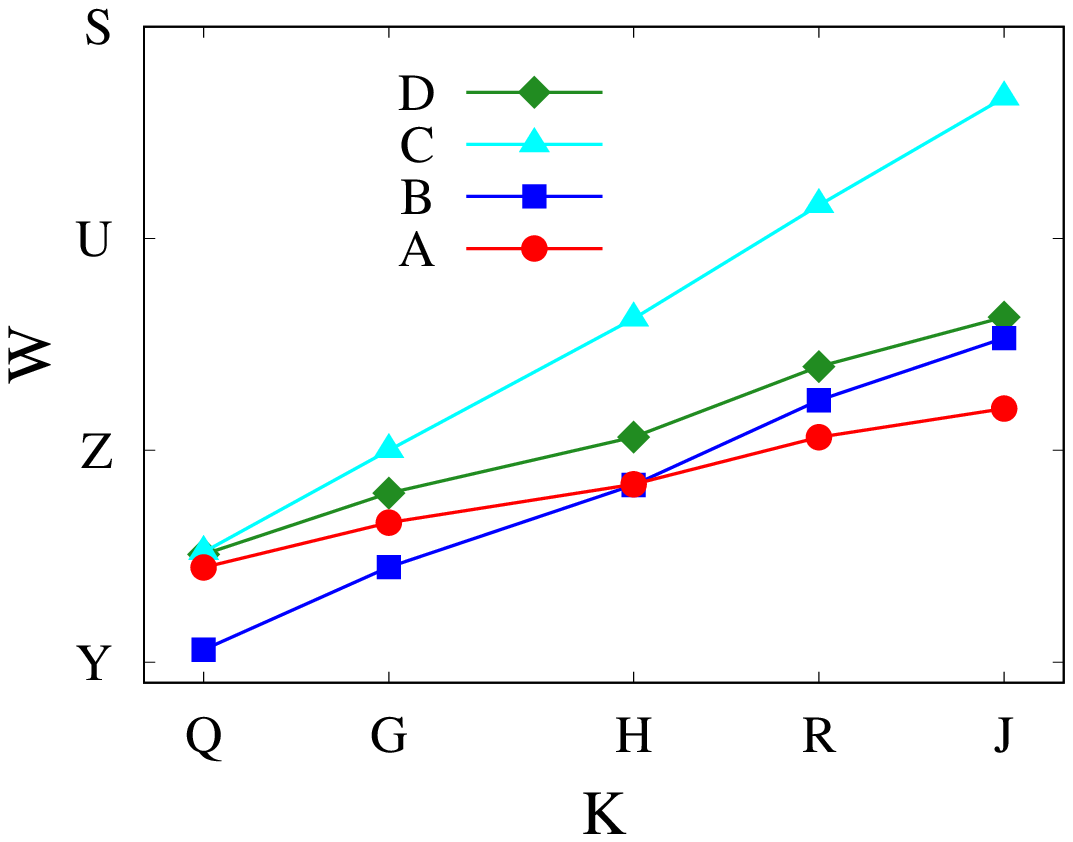}
	}\\ \hspace*{1mm}
	\subfigure[{\normalsize Branch misprediction}]{
		\psfrag{K}[c][c][0.88]{
			\begin{picture}(0,0)
				\put(0,0){\makebox(0,-5)[c]{Number of clusters: $k$ ($\times\! 10^3$)}}
			\end{picture}
		}
		\psfrag{W}[c][c][0.88]{
			\begin{picture}(0,0)
				\put(0,0){\makebox(0,32)[c]{BM}}
			\end{picture}
		}
		\psfrag{Q}[c][c][0.8]{$1$}
		\psfrag{R}[c][c][0.8]{$10$}
		\psfrag{G}[c][c][0.8]{$2$}
		\psfrag{H}[c][c][0.8]{$5$}
		\psfrag{J}[c][c][0.8]{$20$}
		\psfrag{X}[r][r][0.8]{$10^{8}$}
		\psfrag{Y}[r][r][0.8]{$10^{9}$}
		\psfrag{Z}[r][r][0.8]{$10^{10}$}
		\psfrag{U}[r][r][0.8]{$10^{11}$}
		\psfrag{A}[r][r][0.63]{\em SIVF}
		\psfrag{B}[r][r][0.63]{{\em Ding}$^+$}
		\psfrag{C}[r][r][0.63]{{\em Drake}$^+$}
		\psfrag{D}[r][r][0.63]{\em Lloyd-ICP}
		\includegraphics[width=40mm]{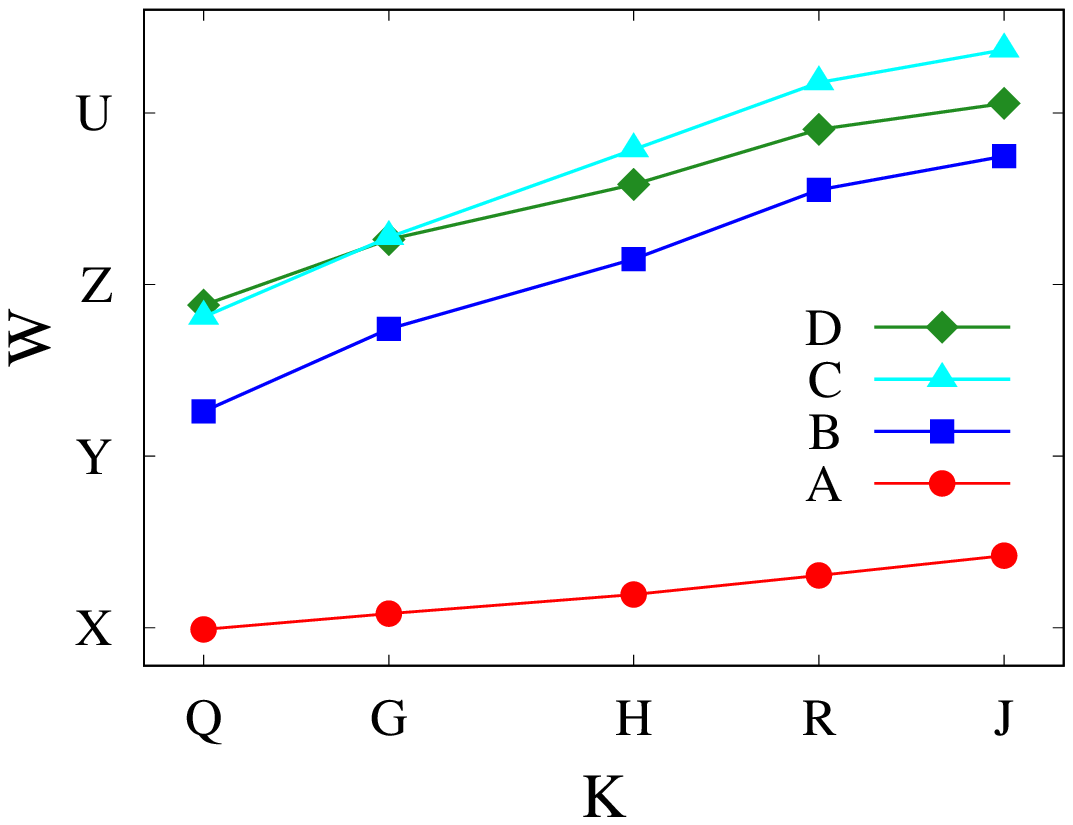}
	}
	\subfigure[{\normalsize LLC miss}]{
		\psfrag{K}[c][c][0.88]{
			\begin{picture}(0,0)
				\put(0,0){\makebox(0,-5)[c]{Number of clusters: $k$ ($\times\! 10^3$)}}
			\end{picture}
		}
		\psfrag{W}[c][c][0.88]{
			\begin{picture}(0,0)
				\put(0,0){\makebox(0,32)[c]{LLCM}}
			\end{picture}
		}
		\psfrag{Q}[c][c][0.8]{$1$}
		\psfrag{R}[c][c][0.8]{$10$}
		\psfrag{G}[c][c][0.8]{$2$}
		\psfrag{H}[c][c][0.8]{$5$}
		\psfrag{J}[c][c][0.8]{$20$}
		\psfrag{X}[r][r][0.8]{$10^8$}
		\psfrag{Y}[r][r][0.8]{$10^9$}
		\psfrag{Z}[r][r][0.8]{$10^{10}$}
		\psfrag{U}[r][r][0.8]{$10^{11}$}
		\psfrag{S}[r][r][0.8]{$10^{12}$}
		\psfrag{A}[r][r][0.63]{\em SIVF}
		\psfrag{B}[r][r][0.63]{\em Ding$^+$}
		\psfrag{C}[r][r][0.63]{\em Drake$^+$}
		\psfrag{D}[r][r][0.63]{\em Lloyd-ICP}
		\includegraphics[width=40mm]{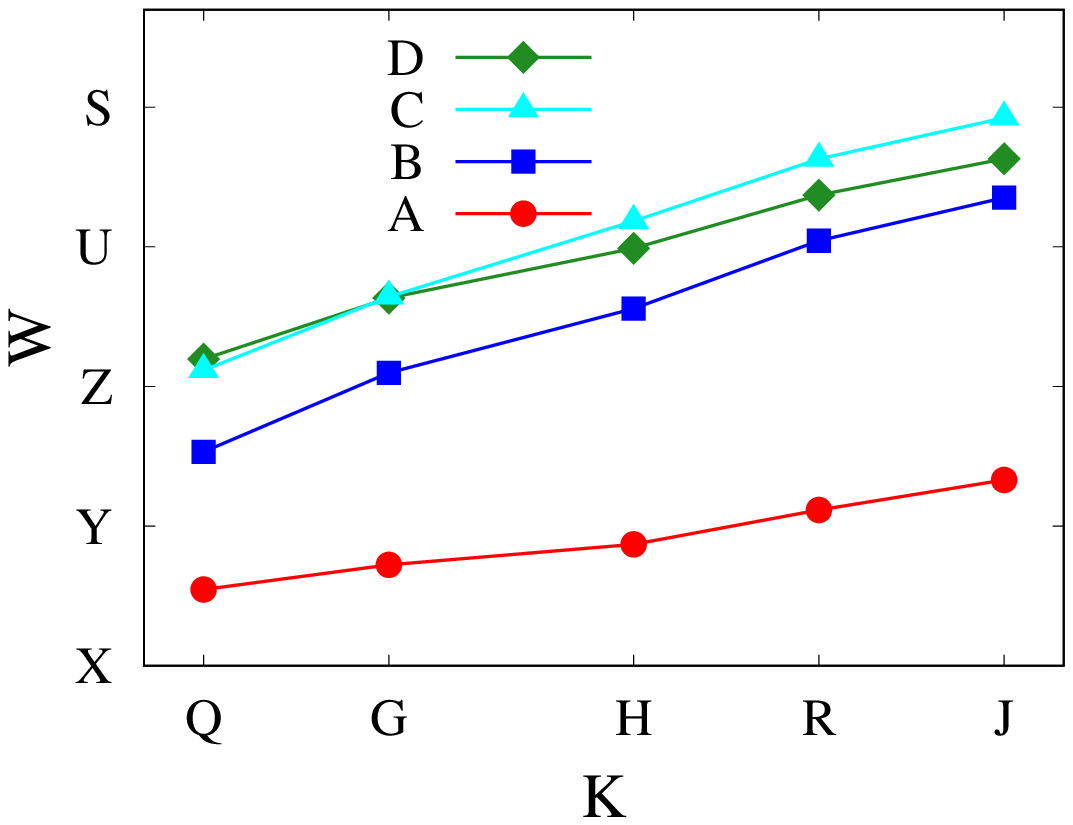}
	}%
\end{center}
\vspace*{-3mm}
\caption{
(a) Average number of similarity calculations normalized by (N x k) 
until the convergence and 
characteristics of performance degradation factors: (b) the number of instructions (Inst), 
(c) branch mispredictions (BM), and (d) last-level cache misses (LLCM) 
when the three algorithms were executed by 50-thread processing for 1M-sized PubMed.
(a) is plotted with log-linear scale and (b), (c), and (d) with log-log scale.
}
\label{fig:comp_dfs_pubmed}
\vspace*{-2mm}
\end{figure}

We demonstrate that the elapsed time that an algorithm needs at run-time 
crucially 
depends on not only the number of expensive similarity calculations 
but also the performance degradation factors (DFs) related to a computer architecture.
Figure~\ref{fig:comp_dfs_pubmed}(a) shows the average number of similarity calculations
per iteration that is normalized by ($N\!\times\!k$) 
when the four algorithms varying $k$ were executed 
by 50-thread processing for 1M-sized PubMed.
It is clear that the Ding's algorithm ({\em Ding}$^+$) remarkably reduced 
the similarity calculations in all $k$ range.
Note that the performances in Fig.~\ref{fig:cal_pubmed} correspond to the points at $k\!=\!20,000$ 
in Fig.~\ref{fig:comp_dfs_pubmed}(a).
Figures~\ref{fig:comp_dfs_pubmed}(b), (c), and (d) show 
the characteristics of DFs, the number of (b) retired (successfully completed) instructions (Inst), 
(c) branch mispredictions (BM), and (d) last-level cache misses (LLCM).
The numbers of retired instructions of the three algorithms except {\em Drake}$^+$ 
were within one order of magnitude in Fig.~\ref{fig:comp_dfs_pubmed}(b).
By contrast, 
BM and LLCM of {\em SIVF} were extremely small, compared with those of the others.
In particular, at $k\!=\!20,000$, the rates of {\em SIVF}'s BM and LLCM to {\em Ding}$^+$'s 
were 0.47\% and 0.95\%, respectively.
Since penalties of a branch misprediction and a last-level cache miss
substantially delay the process 
\cite{evers,eyerman,yasin,ivf}, 
the foregoing differences in DFs have a severe impact on the elapsed time.
In fact,
the penalty of clock cycles becomes several tens to several hundreds times as 
high as the number of clock cycles per retired instruction
in a modern computer system with out-of-order superscalar execution.

Thus architecture-friendly
{\em SIVF} achieved the high-speed performance by suppressing 
the DFs rather than less similarity calculations.
In the following section,
we discuss the effect of giving an appropriate structure to a data set, i.e.,
a structured inverted file for a mean set,
comparing {\em SIVF} with {\em IVF-CBICP} using an unstructured inverted file
in Section~\ref{subsec:cbicp}.

\section{Discussion}\label{sec:disc}
We detail a positive effect of the structured inverted file 
on the elapsed time,  
comparing {\em SIVF} with two prepared algorithms:
One is baseline algorithm {\em IVF} employing only an inverted-file 
for a mean (centroid) set without ICP.
The other is na\"{i}ve algorithm {\em IVF-CBICP} that utilizes ICP 
with a conditional branch shown in Section~\ref{subsec:cbicp}.
Since inverted-file mean array $\breve{\bm{\xi}}_s^{[r-1]}$ in {\em IVF-CBICP} 
has no structure, 
all the centroids $c_{(s,q)}$, $1\!\leq\! q\!\leq\! (mf)_s$, 
are evaluated at the inner-most loop 
in lines~\ref{algo:cbicp_innerloop0} to \ref{algo:cbicp_innerloop1} 
in Algorithm~\ref{algo:cbicp_assign}
whether a similarity calculation between the $c_{(s,q)}$th centroid and the $i$th object
is necessary or not using the conditional branch at line~\ref{algo:cbicp_deepcb} 
in Algorithm~\ref{algo:cbicp_assign}.

\begin{figure}[t]
\begin{center}\hspace*{1mm}
	\subfigure[{\normalsize Avg. elapsed time}]{
		\psfrag{K}[c][c][0.88]{
			\begin{picture}(0,0)
				\put(0,0){\makebox(0,-5)[c]{Number of clusters: $k$ ($\times\! 10^3$)}}
			\end{picture}
		}
		\psfrag{W}[c][c][0.88]{
			\begin{picture}(0,0)
				\put(0,0){\makebox(0,20)[c]{Avg. elapsed time (sec)}}
			\end{picture}
		}
		\psfrag{Q}[c][c][0.8]{$1$}
		\psfrag{R}[c][c][0.8]{$10$}
		\psfrag{G}[c][c][0.8]{$2$}
		\psfrag{H}[c][c][0.8]{$5$}
		\psfrag{J}[c][c][0.8]{$20$}
		\psfrag{X}[r][r][0.8]{$1$}
		\psfrag{Y}[r][r][0.8]{$10$}
		\psfrag{Z}[r][r][0.8]{$10^2$}
		\psfrag{A}[r][r][0.63]{\em SIVF}
		\psfrag{B}[r][r][0.63]{{\em IVF-CBICP}}
		\psfrag{C}[r][r][0.63]{\em IVF}
		\includegraphics[width=40.2mm]{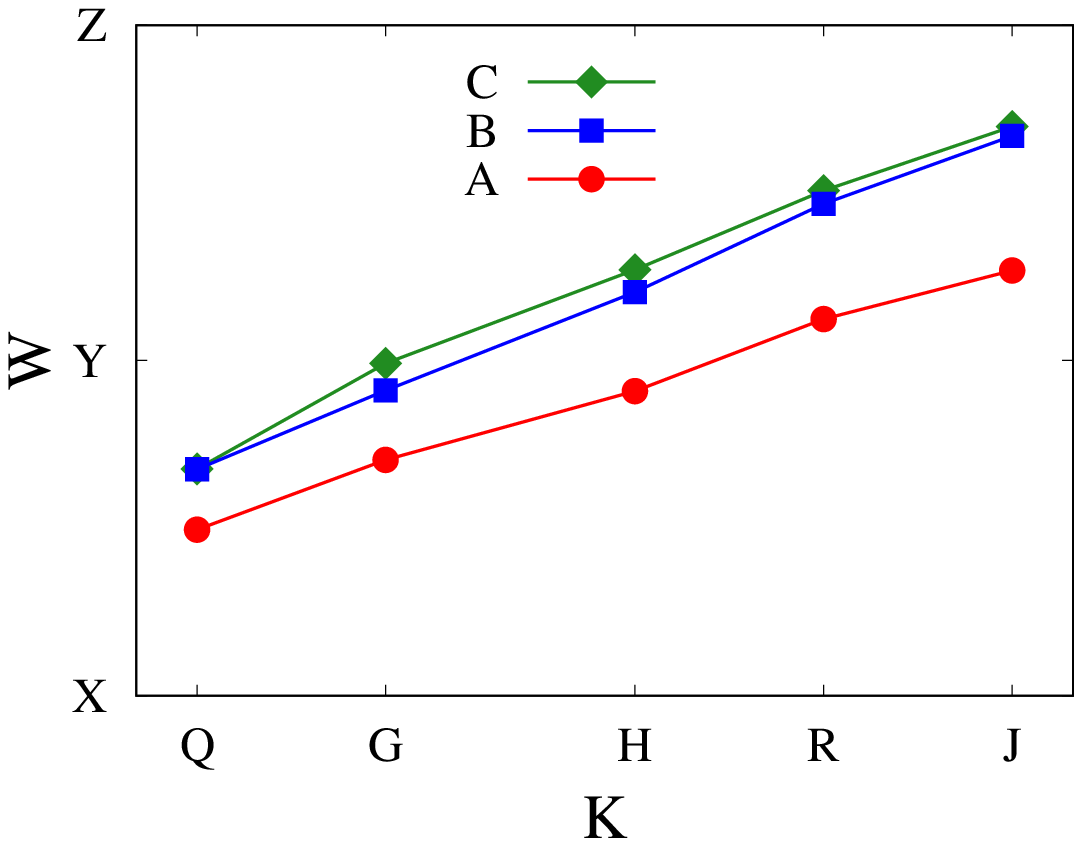}
	}\hspace*{1mm} 
	\subfigure[{\normalsize Max. memory size}]{
		\psfrag{K}[c][c][0.88]{
			\begin{picture}(0,0)
				\put(0,0){\makebox(0,-5)[c]{Number of clusters: $k$ ($\times\! 10^3$)}}
			\end{picture}
		}
		\psfrag{W}[c][c][0.88]{
			\begin{picture}(0,0)
				\put(0,0){\makebox(0,20)[c]{Max. memory size (GB)}}
			\end{picture}
		}
		\psfrag{Q}[c][c][0.8]{$1$}
		\psfrag{R}[c][c][0.8]{$10$}
		\psfrag{G}[c][c][0.8]{$2$}
		\psfrag{H}[c][c][0.8]{$5$}
		\psfrag{J}[c][c][0.8]{$20$}
		\psfrag{X}[r][r][0.8]{$0.6$}
		\psfrag{Y}[r][r][0.8]{$0.8$}
		\psfrag{Z}[r][r][0.8]{$1.0$}
		\psfrag{U}[r][r][0.8]{$1.2$}
		\psfrag{A}[r][r][0.63]{\em SIVF}
		\psfrag{B}[r][r][0.63]{{\em IVF-CBICP}}
		\psfrag{C}[r][r][0.63]{\em IVF}
		\includegraphics[width=40.2mm]{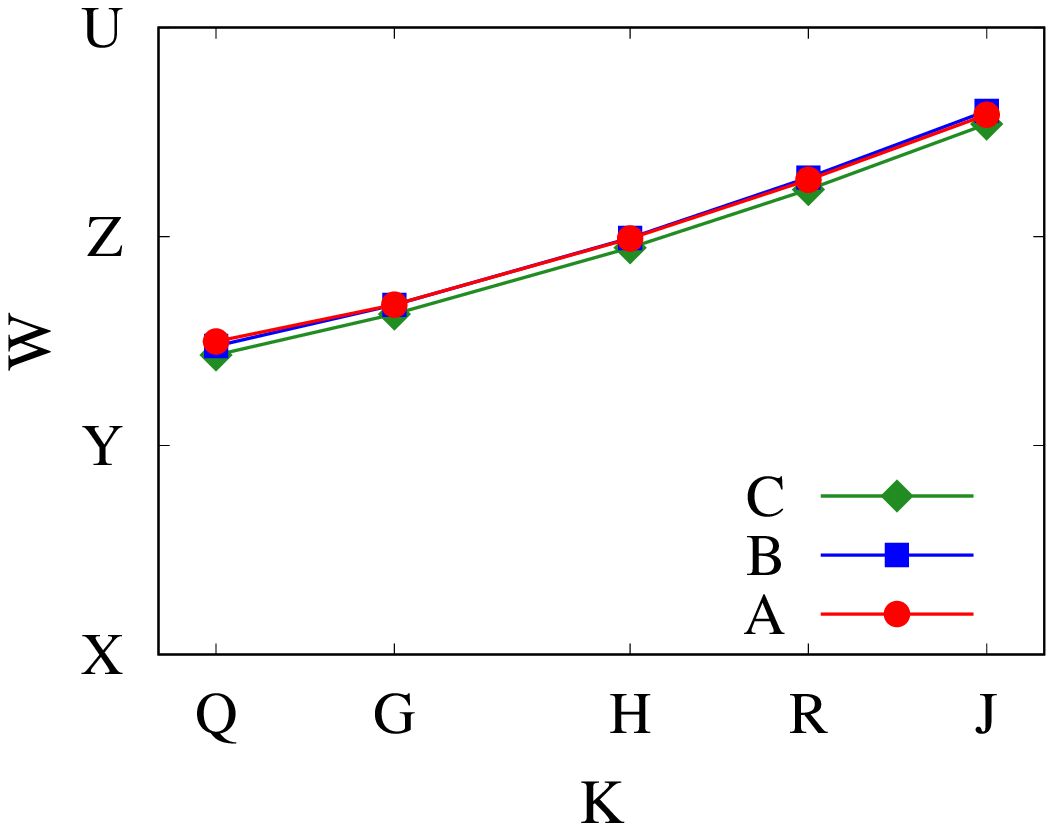}
	}
\end{center}
\vspace*{-3mm}
\caption{Performance of three algorithms that were executed by 50-thread processing 
with given k in 1M-sized PubMed.
(a) Average elapsed time per iteration is plotted along k with log-log scale
and (b) Occupied maximum memory size through iterations along k 
with log-linear scale. 
}
\label{fig:icp_perform}
\vspace*{-2mm}
\end{figure}

Figure~\ref{fig:icp_perform} shows the performance of the three algorithms 
that were executed by 50-thread processing in 1M-sized PubMed, given $k$ values.
The maximum physical memory sizes used by the algorithms were almost the same 
in all the $k$ range in Fig.~\ref{fig:icp_perform}(b) 
because the algorithms did not have much difference in their object and mean data sizes.
In terms of the speed performance in Fig.~\ref{fig:icp_perform}(a), 
{\em SIVF} operated faster than the others 
and the performance difference increased with $k$.
This difference between {\em SIVF} and {\em IVF-CBICP} came from the structure
of the inverted-file arrays.

\begin{figure}[t]
\begin{center}
	\hspace*{1mm}
	\subfigure[{\normalsize Similarity calculation}]{
		\psfrag{K}[c][c][0.88]{
			\begin{picture}(0,0)
				\put(0,0){\makebox(0,-5)[c]{Number of clusters: $k$ ($\times\! 10^3$)}}
			\end{picture}
		}
		\psfrag{W}[c][c][0.88]{
			\begin{picture}(0,0)
				\put(0,0){\makebox(0,22)[c]{\# similarity cal.}}
			\end{picture}
		}
		\psfrag{Q}[c][c][0.8]{$1$}
		\psfrag{R}[c][c][0.8]{$10$}
		\psfrag{G}[c][c][0.8]{$2$}
		\psfrag{H}[c][c][0.8]{$5$}
		\psfrag{J}[c][c][0.8]{$20$}
		\psfrag{X}[r][r][0.8]{$0$}
		\psfrag{Y}[r][r][0.8]{$0.2$}
		\psfrag{Z}[r][r][0.8]{$0.4$}
		\psfrag{U}[r][r][0.8]{$0.6$}
		\psfrag{S}[r][r][0.8]{$0.8$}
		\psfrag{T}[r][r][0.8]{$1.0$}
		\psfrag{A}[r][r][0.63]{\em SIVF}
		\psfrag{B}[r][r][0.63]{{\em IVF-CBICP}}
		\psfrag{C}[r][r][0.63]{\em IVF}
		\includegraphics[width=40mm]{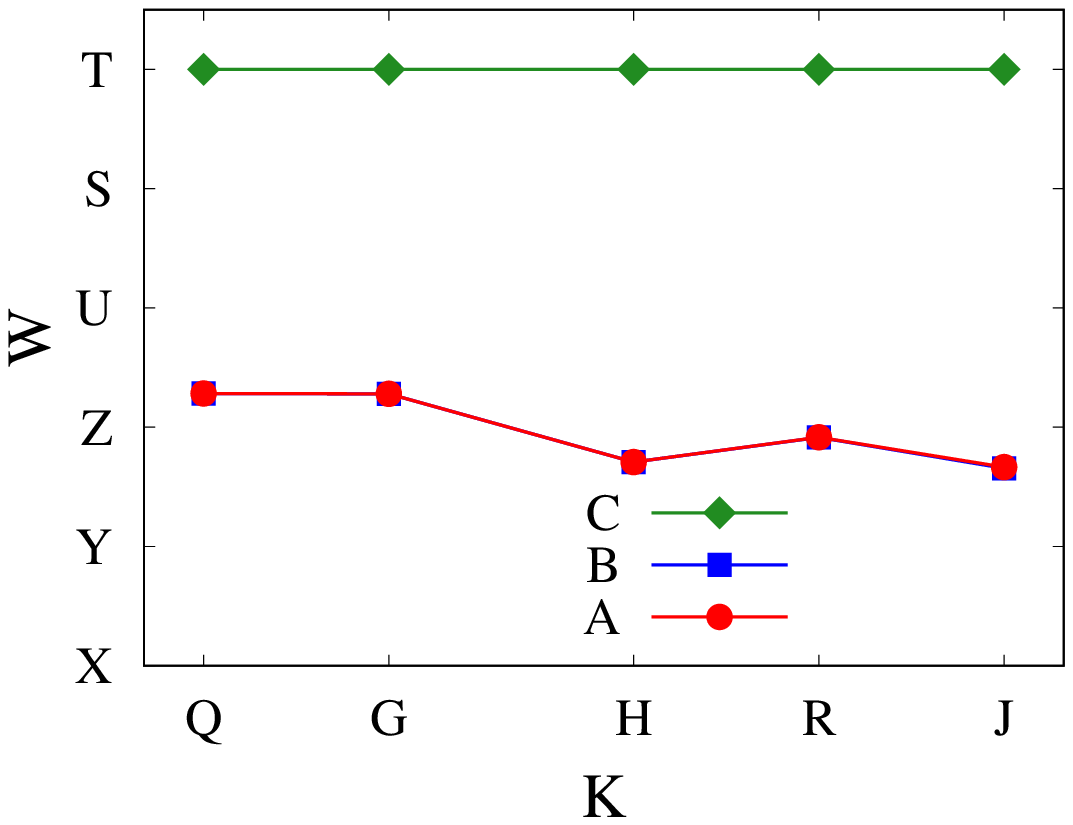}
	}\hspace*{1mm}
	\subfigure[{\normalsize Instruction}]{
		\psfrag{K}[c][c][0.88]{
			\begin{picture}(0,0)
				\put(0,0){\makebox(0,-5)[c]{Number of clusters: $k$ ($\times\! 10^3$)}}
			\end{picture}
		}
		\psfrag{W}[c][c][0.88]{
			\begin{picture}(0,0)
				\put(0,0){\makebox(0,32)[c]{\# instructions}}
			\end{picture}
		}
		\psfrag{Q}[c][c][0.8]{$1$}
		\psfrag{R}[c][c][0.8]{$10$}
		\psfrag{G}[c][c][0.8]{$2$}
		\psfrag{H}[c][c][0.8]{$5$}
		\psfrag{J}[c][c][0.8]{$20$}
		\psfrag{X}[r][r][0.8]{$10^{11}$}
		\psfrag{Y}[r][r][0.8]{$10^{12}$}
		\psfrag{Z}[r][r][0.8]{$10^{13}$}
		\psfrag{A}[r][r][0.63]{\em SIVF}
		\psfrag{B}[r][r][0.63]{{\em IVF-CBICP}}
		\psfrag{C}[r][r][0.63]{\em IVF}
		\includegraphics[width=40mm]{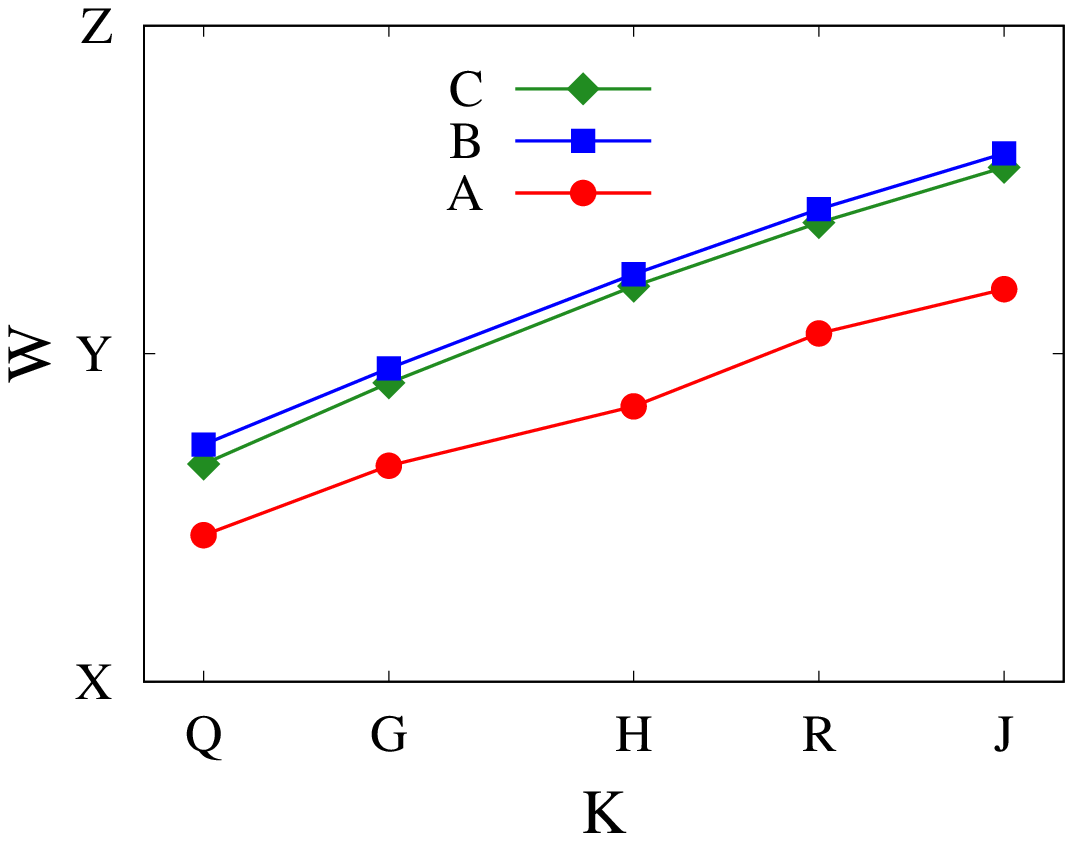}
	}\\ \hspace*{1mm}
	\subfigure[{\normalsize Branch misprediction}]{
		\psfrag{K}[c][c][0.88]{
			\begin{picture}(0,0)
				\put(0,0){\makebox(0,-5)[c]{Number of clusters: $k$ ($\times\! 10^3$)}}
			\end{picture}
		}
		\psfrag{W}[c][c][0.88]{
			\begin{picture}(0,0)
				\put(0,0){\makebox(0,22)[c]{BM}}
			\end{picture}
		}
		\psfrag{Q}[c][c][0.8]{$1$}
		\psfrag{R}[c][c][0.8]{$10$}
		\psfrag{G}[c][c][0.8]{$2$}
		\psfrag{H}[c][c][0.8]{$5$}
		\psfrag{J}[c][c][0.8]{$20$}
		\psfrag{X}[r][r][0.8]{$10^{7}$}
		\psfrag{Y}[r][r][0.8]{$10^{8}$}
		\psfrag{Z}[r][r][0.8]{$10^{9}$}
		\psfrag{U}[r][r][0.8]{$10^{10}$}
		\psfrag{A}[r][r][0.63]{\em SIVF}
		\psfrag{B}[r][r][0.63]{{\em IVF-CBICP}}
		\psfrag{C}[r][r][0.63]{\em IVF}
		\includegraphics[width=40mm]{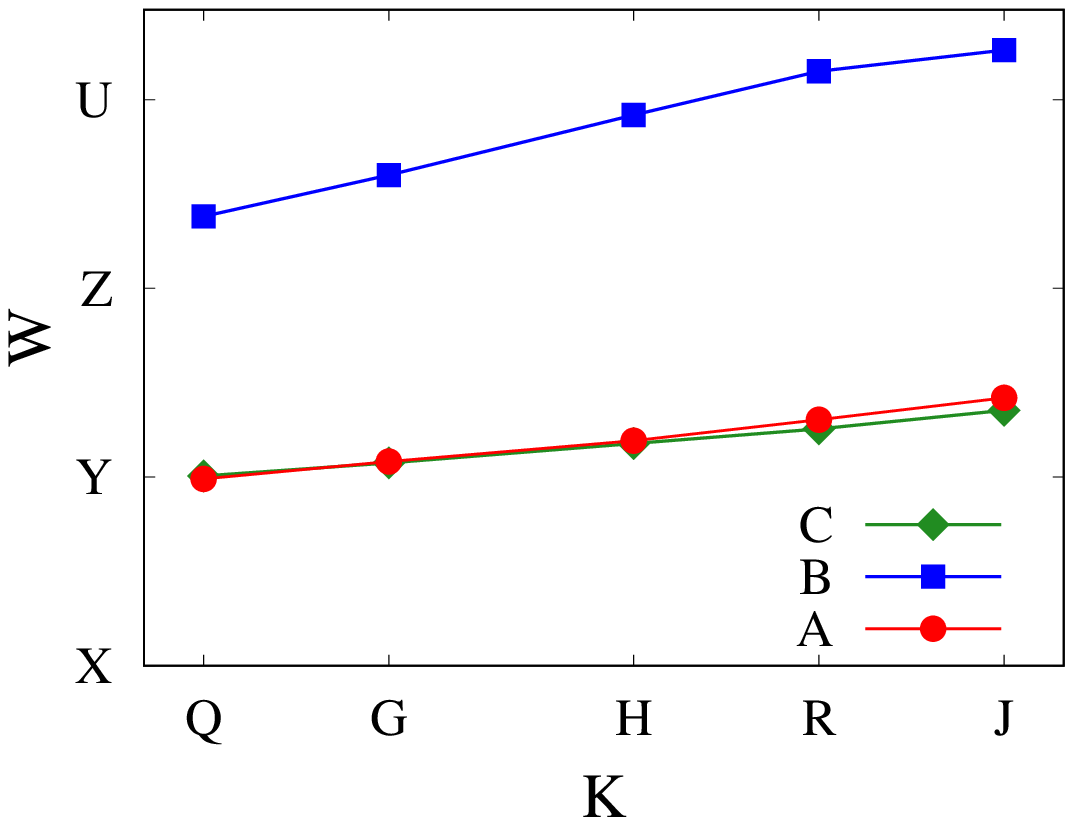}
	}
	\subfigure[{\normalsize LLC miss}]{
		\psfrag{K}[c][c][0.88]{
			\begin{picture}(0,0)
				\put(0,0){\makebox(0,-5)[c]{Number of clusters: $k$ ($\times\! 10^3$)}}
			\end{picture}
		}
		\psfrag{W}[c][c][0.88]{
			\begin{picture}(0,0)
				\put(0,0){\makebox(0,22)[c]{LLCM}}
			\end{picture}
		}
		\psfrag{Q}[c][c][0.8]{$1$}
		\psfrag{R}[c][c][0.8]{$10$}
		\psfrag{G}[c][c][0.8]{$2$}
		\psfrag{H}[c][c][0.8]{$5$}
		\psfrag{J}[c][c][0.8]{$20$}
		\psfrag{X}[r][r][0.8]{$10^8$}
		\psfrag{Y}[r][r][0.8]{$10^9$}
		\psfrag{Z}[r][r][0.8]{$10^{10}$}
		\psfrag{A}[r][r][0.63]{\em SIVF}
		\psfrag{B}[r][r][0.63]{{\em IVF-CBICP}}
		\psfrag{C}[r][r][0.63]{\em IVF}
		\includegraphics[width=40mm]{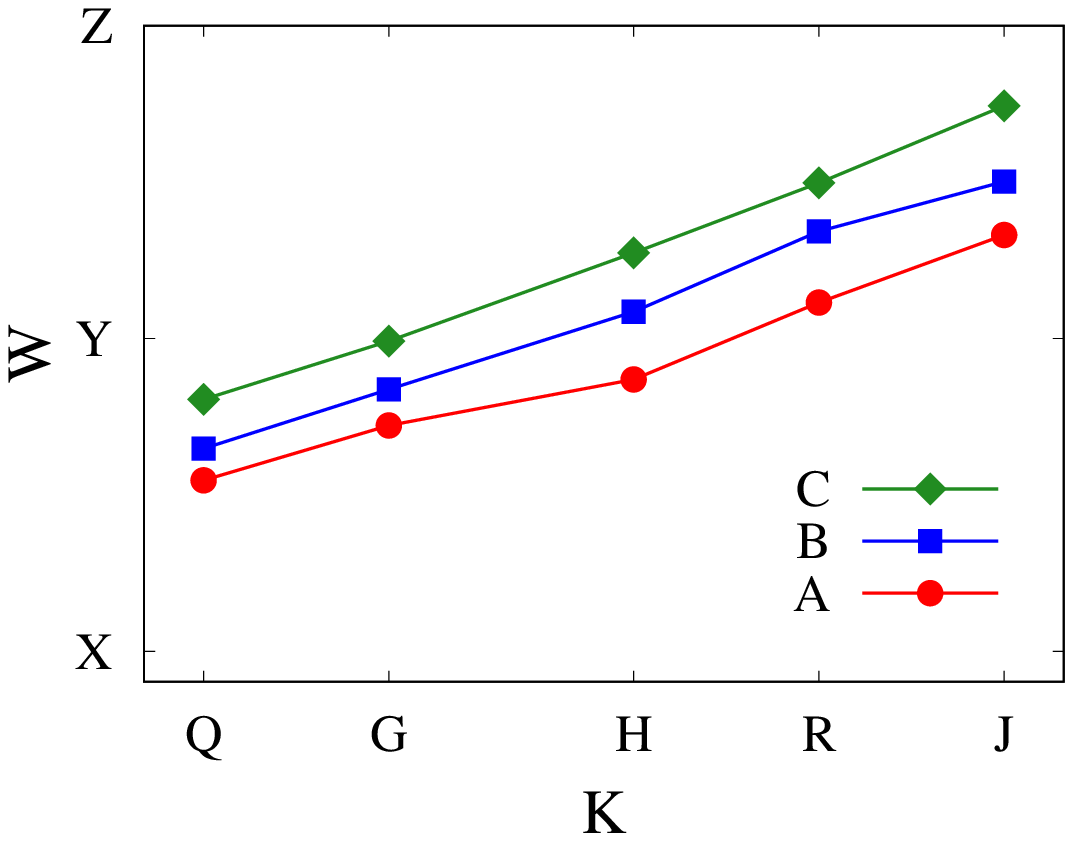}
	}%
\end{center}
\vspace*{-3mm}
\caption{
(a) Number of similarity calculations normalized by (N x k) and 
characteristics of performance degradation factors: (b) the number of instructions (Inst), 
(c) branch mispredictions (BM), and (d) last-level cache misses (LLCM) 
when the three algorithms were executed by 50-thread processing for 1M-sized PubMed.
(a) is plotted with log-linear scale and (b), (c), and (d) with log-log scale.
}
\label{fig:icp_dfs_pubmed}
\vspace*{-2mm}
\end{figure}

We first observe the number of similarity calculations 
before discussing the effects of the DFs.
Figure~\ref{fig:icp_dfs_pubmed}(a) shows that
the number of similarity calculations normalized by ($N\!\times\!k$).
The baseline algorithm, {\em IVF} without ICP, performed all the similarity calculations 
like Lloyd's algorithm.
By contrast, 
{\em SIVF} and {\em IVF-CBICP} 
executed the same number of the similarity calculations, which corresponded to 
only 30\% to 65\% of the baseline.
However, there was the large difference between the elapsed times of 
{\em SIVF} and {\em IVF-CBICP} in Fig.~\ref{fig:icp_perform}(a).

We analyzed the {\em SIVF} speed performance from the viewpoint of DFs.
Figure~\ref{fig:icp_dfs_pubmed}(b), (c), and (d) show 
the characteristics of the DFs; 
the number of retired instructions, BMs, and LLCMs.
The number of retired instructions of {\em SIVF} was smallest 
in all the $k$ values because of the reduction of instructions 
for similarity calculations with ICP in Fig.~\ref{fig:icp_dfs_pubmed}(a).
In the case of {\em IVF-CBICP}, positive and negative effects 
on the number of instructions compensated
by decreasing the number of similarity calculations and 
increasing the number of conditional branches,
resulting in the similar characteristics as {\em IVF}.
In Fig.~\ref{fig:icp_dfs_pubmed}(c),
{\em SIVF} reduced the branch mispredictions as much as {\em IVF}
although {\em IVF-CBICP} caused many branch misprediction.
{\em SIVF} also reduced the last-level cache misses by skipping unnecessary 
similarity calculations with ICP based on the structured inverted file.

Thus suppressing the DFs as well as reducing the similarity calculations,
i.e., exploiting the advantages of the computer architecture,
led to the {\em SIVF}'s high-speed performance.

\section{Conclusion}\label{sec:conc}
We proposed an architecture-friendly
structured inverted-file $k$-means clustering algorithm 
({\em SIVF})
that operated at higher speed and with lower memory consumption 
in large-scale high-dimensional sparse document data sets 
when large $k$ values were given, 
compared with the existing algorithms.
Our analysis on the experimental results revealed that
{\em SIVF}'s high-performance came from suppressing 
the performance degradation factors of
the numbers of cache misses and branch mispredictions 
rather than decreasing the number of expensive similarity calculations.
Our approach of devising a data structure to exploit advantages of
computer architecture
provides an algorithm design guideline for 
large-scale and high-dimensional sparse data sets.

There remain the two directions as the future work.
One is to clarify the limitations of our algorithm, 
for instance, on the parameters of $k$, data size $N$, the sparsity 
of an object data set and a mean set, and the characteristics of the 
object data set like the power-law distribution of appearing terms.
The other is to develop more efficient filter that
can reduce more similarity calculations instead of the weak ICP and 
incorporate it into {\em SIVF} 
so as to become an architecture-friendly algorithm.


\newpage
\appendices
\section{Performance Comparison Results in NYT}\label{sec:appA}
\begin{figure}[ht]
\begin{center}\hspace*{1mm}
	\begin{tabular}{cc}	
	\subfigure{
		\psfrag{K}[c][c][0.88]{
			\begin{picture}(0,0)
				\put(0,0){\makebox(0,-5)[c]{Number of clusters: $k$ ($\times\! 10^3$)}}
			\end{picture}
		}
		\psfrag{T}[c][c][0.88]{
			\begin{picture}(0,0)
				\put(0,0){\makebox(0,20)[c]{Avg. elapsed time (sec)}}
			\end{picture}
		}
		\psfrag{Q}[c][c][0.8]{$1$}
		\psfrag{R}[c][c][0.8]{$10$}
		\psfrag{S}[c][c][0.8]{$2$}
		\psfrag{W}[c][c][0.8]{$5$}
		\psfrag{V}[c][c][0.8]{$20$}
		\psfrag{X}[r][r][0.8]{$1$}
		\psfrag{Y}[r][r][0.8]{$10$}
		\psfrag{Z}[r][r][0.8]{$10^2$}
		\psfrag{U}[r][r][0.8]{$10^3$}
		\psfrag{A}[r][r][0.63]{\em SIVF}
		\psfrag{B}[r][r][0.63]{{\em Ding}$^+$}
		\psfrag{C}[r][r][0.63]{{\em Drake}$^+$}
		\psfrag{D}[r][r][0.63]{\em Lloyd-ICP}
		\includegraphics[width=40mm]{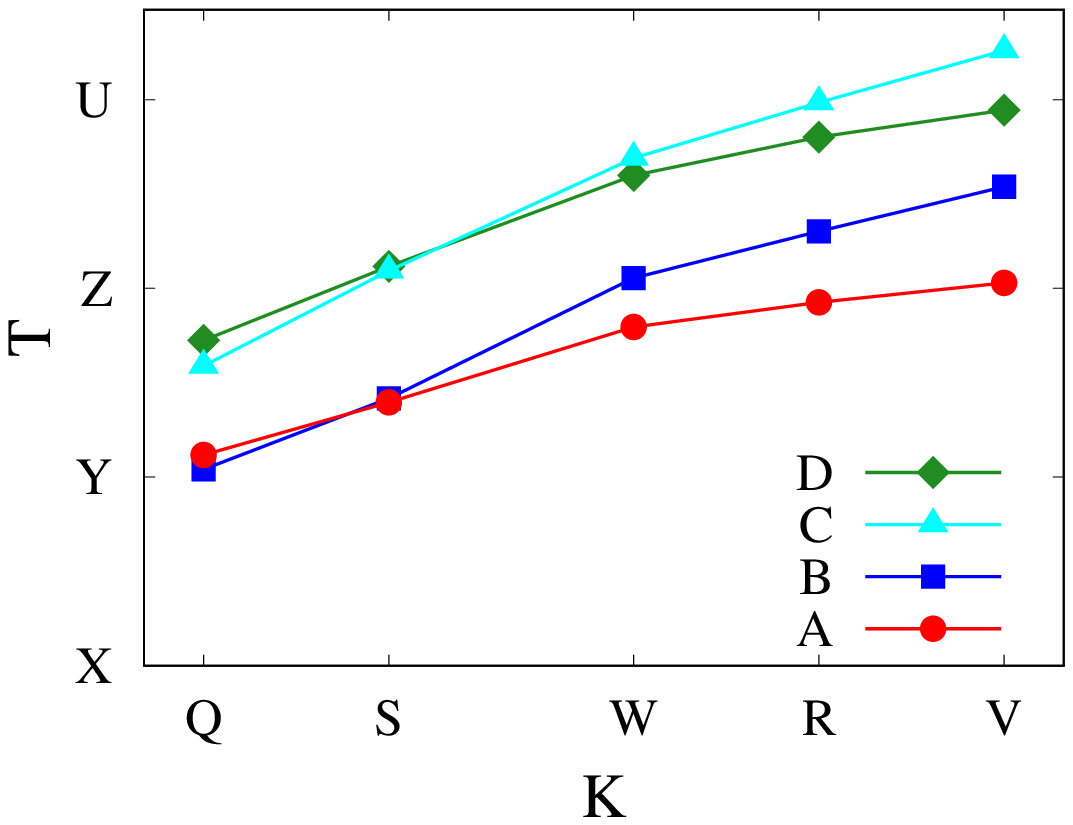}
	} &
	\subfigure{
		\psfrag{K}[c][c][0.88]{
			\begin{picture}(0,0)
				\put(0,0){\makebox(0,-5)[c]{Number of clusters: $k$ ($\times\! 10^3$)}}
			\end{picture}
		}
		\psfrag{T}[c][c][0.88]{
			\begin{picture}(0,0)
				\put(0,0){\makebox(0,20)[c]{Max. memory size (GB)}}
			\end{picture}
		}
		\psfrag{Q}[c][c][0.8]{$1$}
		\psfrag{R}[c][c][0.8]{$10$}
		\psfrag{S}[c][c][0.8]{$2$}
		\psfrag{W}[c][c][0.8]{$5$}
		\psfrag{V}[c][c][0.8]{$20$}
		\psfrag{X}[r][r][0.8]{$1$}
		\psfrag{Y}[r][r][0.8]{$10$}
		\psfrag{Z}[r][r][0.8]{$10^2$}
		\psfrag{A}[r][r][0.63]{\em SIVF}
		\psfrag{B}[r][r][0.63]{{\em Ding}$^+$}
		\psfrag{C}[r][r][0.63]{{\em Drake}$^+$}
		\psfrag{D}[r][r][0.63]{\em Lloyd-ICP}
		\includegraphics[width=40mm]{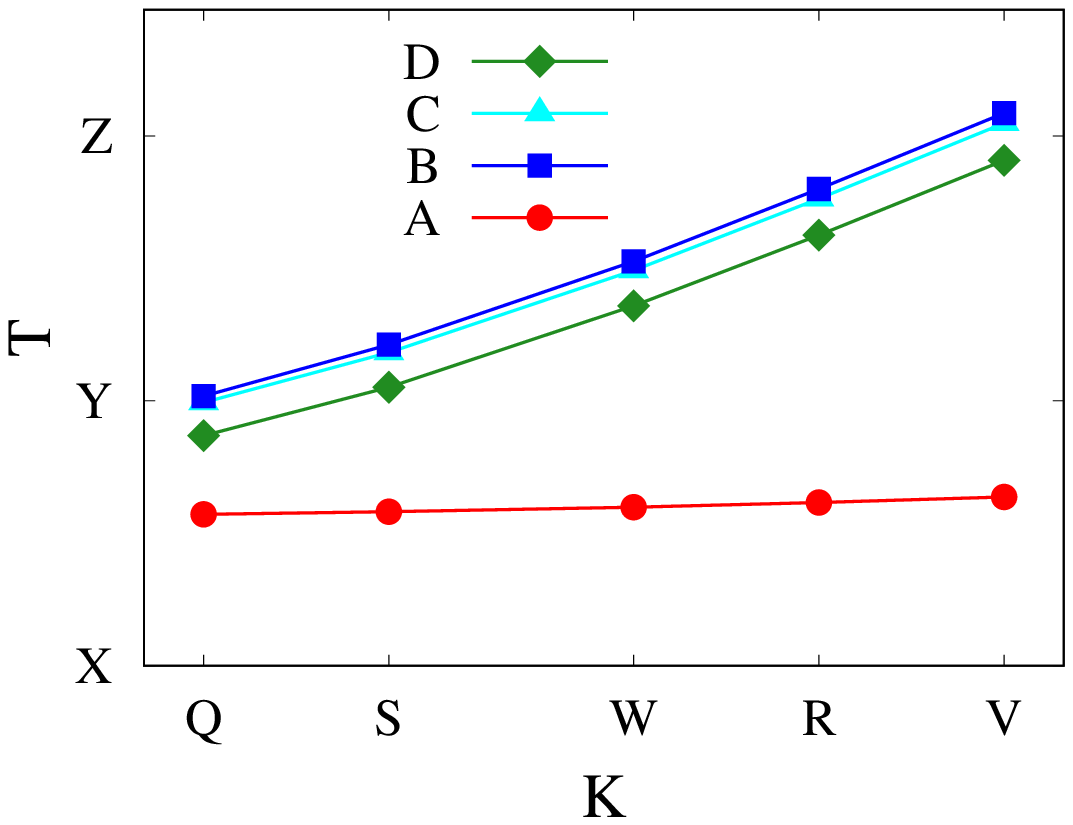}
	}\tabularnewline
	(a)~Avg. elapsed time & (b)~Max. memory size \tabularnewline
	\end{tabular}
\end{center}
\vspace*{-3mm}
\caption{Performance of four algorithms executed by 50-thread processing 
with given k in NYT.
(a) Average elapsed time per iteration and 
(b) Occupied maximum memory size through iterations were plotted along k 
with log-log scale. 
}
\label{fig:perform_nyt}
\vspace*{-2mm}
\end{figure}

\begin{figure}[ht]
\begin{center}
	\psfrag{K}[c][c][0.9]{Iteration}
	\psfrag{T}[c][c][0.9]{
		\begin{picture}(0,0)
			\put(0,0){\makebox(0,22)[c]{Elapsed time (sec)}}
		\end{picture}
	}
	\psfrag{P}[c][c][0.86]{$0$}
	\psfrag{Q}[c][c][0.86]{$20$}
	\psfrag{R}[c][c][0.86]{$40$}
	\psfrag{S}[c][c][0.86]{$60$}
	\psfrag{H}[c][c][0.86]{$80$}
	\psfrag{X}[r][r][0.86]{$10^2$}
	\psfrag{Y}[r][r][0.86]{$10^3$}
	\psfrag{Z}[r][r][0.86]{$10^4$}
	\psfrag{U}[r][r][0.86]{$10^5$}
	\psfrag{A}[r][r][0.7]{\em SIVF}
	\psfrag{B}[r][r][0.7]{{\em Ding}$^+$}
	\psfrag{C}[r][r][0.7]{{\em Drake}$^+$}
	\psfrag{D}[r][r][0.7]{\em Lloyd-ICP}
	\includegraphics[width=50mm]{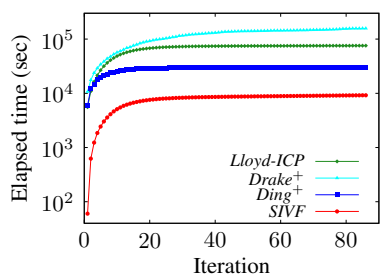}
\end{center}
\vspace*{-3mm}
\caption{
Elapsed time that four algorithms with k=20,000 required until convergence 
when they were applied to NYT.
Elapsed time is plotted along iteration with linear-log scale.
}
\label{fig:itrTime_nyt}
\vspace*{-2mm}
\end{figure}

\begin{figure}[ht]
\begin{center}
	\psfrag{K}[c][c][0.95]{Iteration}
	\psfrag{W}[c][c][0.95]{
		\begin{picture}(0,0)
			\put(0,0){\makebox(0,30)[c]{Normalized \# sim. calc.}}
		\end{picture}
	}
	\psfrag{P}[c][c][0.86]{$0$}
	\psfrag{Q}[c][c][0.86]{$20$}
	\psfrag{R}[c][c][0.86]{$40$}
	\psfrag{S}[c][c][0.86]{$60$}
	\psfrag{T}[c][c][0.86]{$80$}
	\psfrag{X}[r][r][0.86]{$10^{-5}$}
	\psfrag{Y}[r][r][0.86]{$10^{-4}$}
	\psfrag{Z}[r][r][0.86]{$10^{-3}$}
	\psfrag{H}[r][r][0.86]{$10^{-2}$}
	\psfrag{J}[r][r][0.86]{$10^{-1}$}
	\psfrag{L}[r][r][0.86]{$1$}
	\psfrag{A}[r][r][0.7]{\em SIVF}
	\psfrag{B}[r][r][0.7]{{\em Ding}$^+$}
	\psfrag{C}[r][r][0.7]{{\em Drake}$^+$}
	\psfrag{D}[r][r][0.7]{\em Lloyd-ICP}
	\includegraphics[width=53mm]{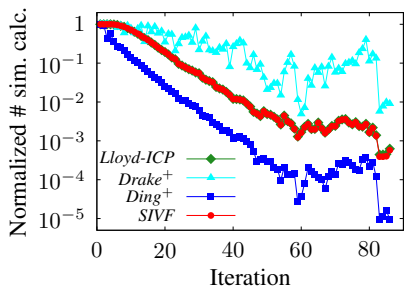}
\end{center}
\vspace*{-3mm}
\caption{
Number of similarity calculations normalized by (N x k), 
where N indicates the data size of 1,285,944, 
when four algorithms with k=20,000 were executed by 50-thread processing for NYT.
Normalized number is plotted along iteration with linear-log scale.
}
\label{fig:mult_nyt}
\vspace*{-2mm}
\end{figure}

\begin{figure}[ht]
\begin{center}
	\hspace*{1mm}
	\subfigure[{\normalsize Similarity calculation}]{
		\psfrag{K}[c][c][0.88]{
			\begin{picture}(0,0)
				\put(0,0){\makebox(0,-5)[c]{Number of clusters: $k$ ($\times\! 10^3$)}}
			\end{picture}
		}
		\psfrag{W}[c][c][0.88]{
			\begin{picture}(0,0)
				\put(0,0){\makebox(0,22)[c]{Normalized \# sim. cal.}}
			\end{picture}
		}
		\psfrag{Q}[c][c][0.8]{$1$}
		\psfrag{R}[c][c][0.8]{$10$}
		\psfrag{G}[c][c][0.8]{$2$}
		\psfrag{H}[c][c][0.8]{$5$}
		\psfrag{J}[c][c][0.8]{$20$}
		\psfrag{X}[r][r][0.8]{$0$}
		\psfrag{Y}[r][r][0.8]{$0.2$}
		\psfrag{Z}[r][r][0.8]{$0.4$}
		\psfrag{U}[r][r][0.8]{$0.6$}
		\psfrag{S}[r][r][0.8]{$0.8$}
		\psfrag{T}[r][r][0.8]{$1.0$}
		\psfrag{A}[r][r][0.63]{\em SIVF}
		\psfrag{B}[r][r][0.63]{{\em Ding}$^+$}
		\psfrag{C}[r][r][0.63]{{\em Drake}$^+$}
		\psfrag{D}[r][r][0.63]{\em Lloyd-ICP}
		\includegraphics[width=40mm]{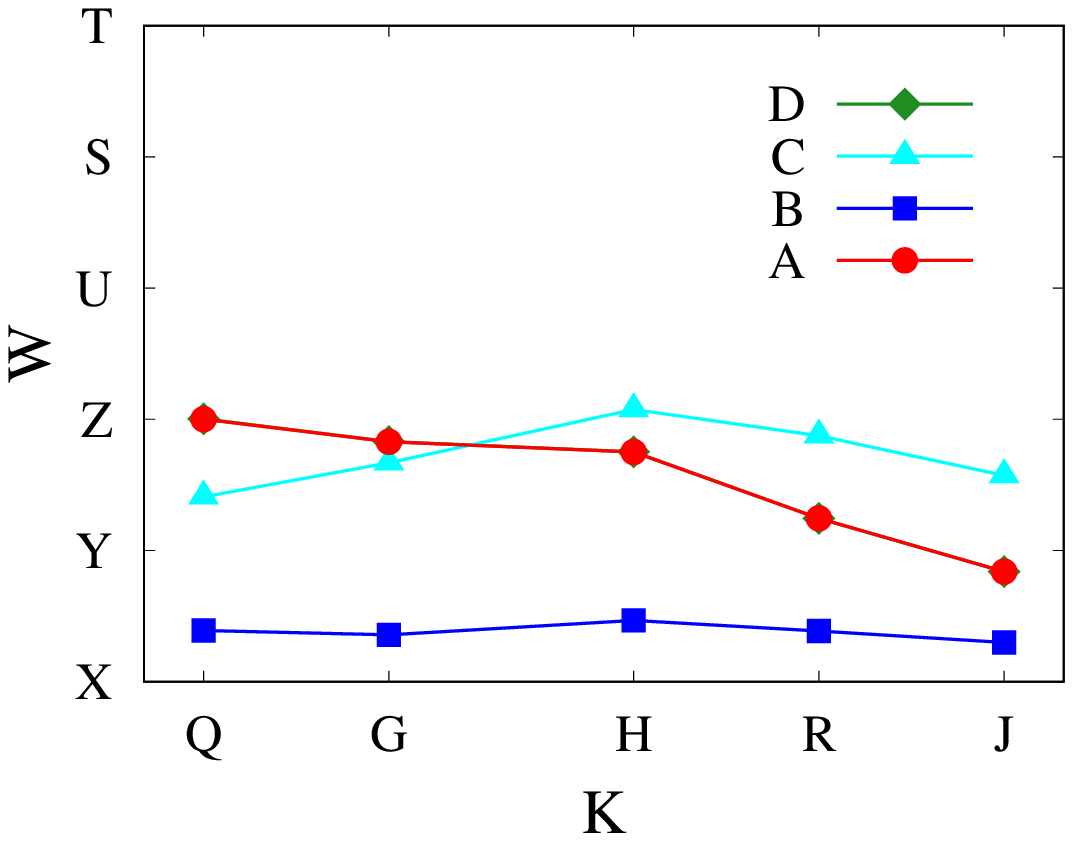}
	}\hspace*{1mm}
	\subfigure[{\normalsize Instruction}]{
		\psfrag{K}[c][c][0.88]{
			\begin{picture}(0,0)
				\put(0,0){\makebox(0,-5)[c]{Number of clusters: $k$ ($\times\! 10^3$)}}
			\end{picture}
		}
		\psfrag{W}[c][c][0.88]{
			\begin{picture}(0,0)
				\put(0,0){\makebox(0,32)[c]{\# instructions}}
			\end{picture}
		}
		\psfrag{Q}[c][c][0.8]{$1$}
		\psfrag{R}[c][c][0.8]{$10$}
		\psfrag{G}[c][c][0.8]{$2$}
		\psfrag{H}[c][c][0.8]{$5$}
		\psfrag{J}[c][c][0.8]{$20$}
		\psfrag{X}[r][r][0.8]{$10^{10}$}
		\psfrag{Y}[r][r][0.8]{$10^{11}$}
		\psfrag{Z}[r][r][0.8]{$10^{12}$}
		\psfrag{U}[r][r][0.8]{$10^{13}$}
		\psfrag{S}[r][r][0.8]{$10^{14}$}
		\psfrag{A}[r][r][0.63]{\em SIVF}
		\psfrag{B}[r][r][0.63]{{\em Ding}$^+$}
		\psfrag{C}[r][r][0.63]{{\em Drake}$^+$}
		\psfrag{D}[r][r][0.63]{\em Lloyd-ICP}
		\includegraphics[width=40mm]{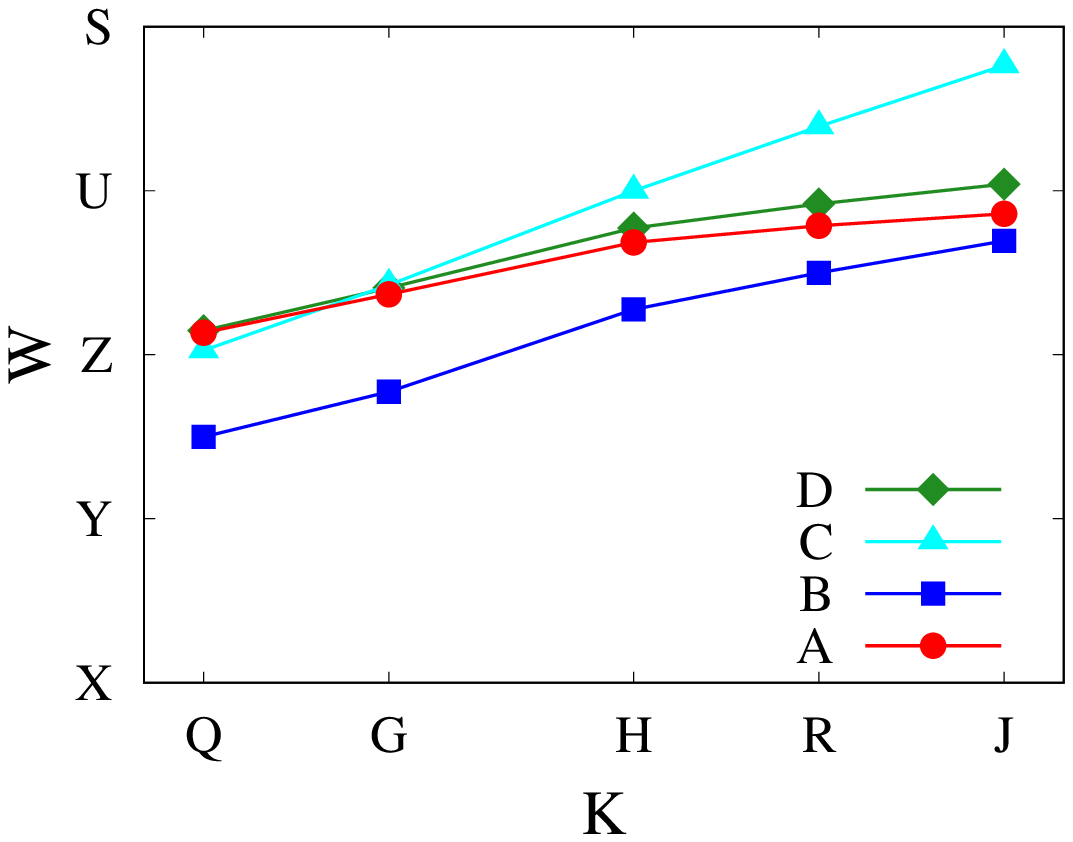}
	}\\ \hspace*{1mm}
	\subfigure[{\normalsize Branch misprediction}]{
		\psfrag{K}[c][c][0.88]{
			\begin{picture}(0,0)
				\put(0,0){\makebox(0,-5)[c]{Number of clusters: $k$ ($\times\! 10^3$)}}
			\end{picture}
		}
		\psfrag{W}[c][c][0.88]{
			\begin{picture}(0,0)
				\put(0,0){\makebox(0,32)[c]{BM}}
			\end{picture}
		}
		\psfrag{Q}[c][c][0.8]{$1$}
		\psfrag{R}[c][c][0.8]{$10$}
		\psfrag{G}[c][c][0.8]{$2$}
		\psfrag{H}[c][c][0.8]{$5$}
		\psfrag{J}[c][c][0.8]{$20$}
		\psfrag{X}[r][r][0.8]{$10^{8}$}
		\psfrag{Y}[r][r][0.8]{$10^{9}$}
		\psfrag{Z}[r][r][0.8]{$10^{10}$}
		\psfrag{U}[r][r][0.8]{$10^{11}$}
		\psfrag{S}[r][r][0.8]{$10^{12}$}
		\psfrag{A}[r][r][0.63]{\em SIVF}
		\psfrag{B}[r][r][0.63]{{\em Ding}$^+$}
		\psfrag{C}[r][r][0.63]{{\em Drake}$^+$}
		\psfrag{D}[r][r][0.63]{\em Lloyd-ICP}
		\includegraphics[width=40mm]{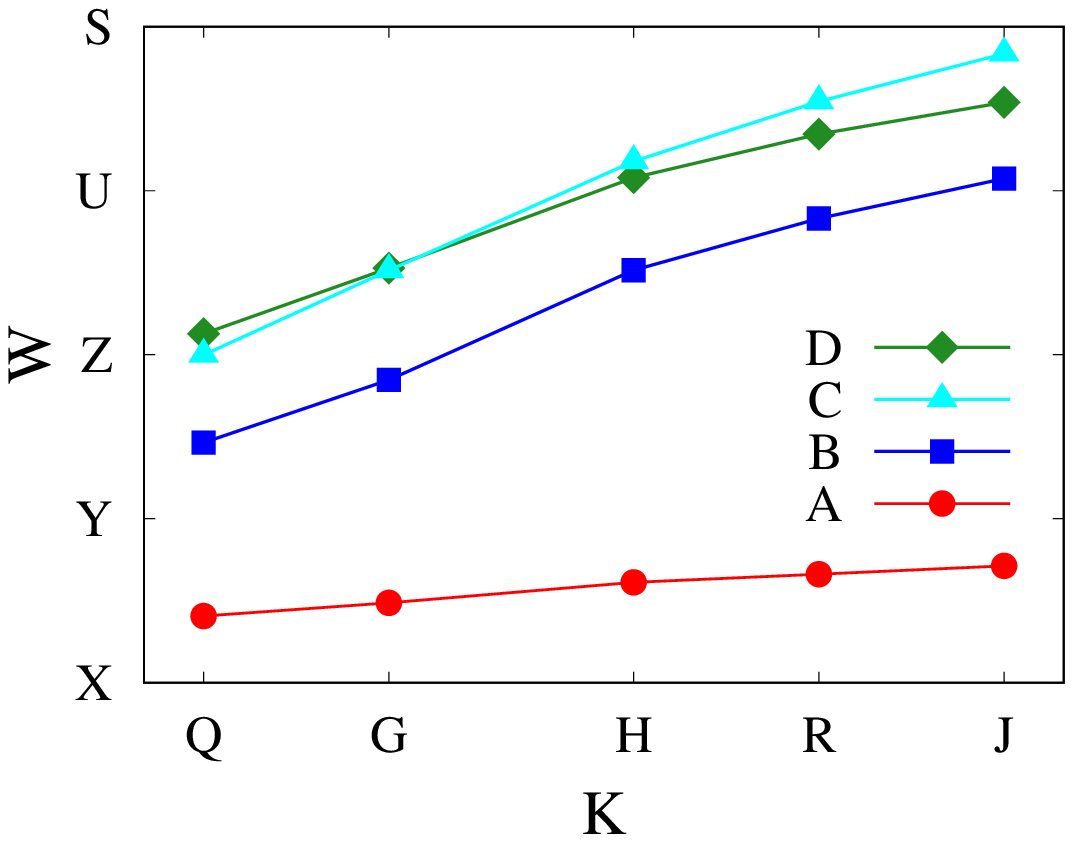}
	}
	\subfigure[{\normalsize LLC miss}]{
		\psfrag{K}[c][c][0.88]{
			\begin{picture}(0,0)
				\put(0,0){\makebox(0,-5)[c]{Number of clusters: $k$ ($\times\! 10^3$)}}
			\end{picture}
		}
		\psfrag{W}[c][c][0.88]{
			\begin{picture}(0,0)
				\put(0,0){\makebox(0,32)[c]{LLCM}}
			\end{picture}
		}
		\psfrag{Q}[c][c][0.8]{$1$}
		\psfrag{R}[c][c][0.8]{$10$}
		\psfrag{G}[c][c][0.8]{$2$}
		\psfrag{H}[c][c][0.8]{$5$}
		\psfrag{J}[c][c][0.8]{$20$}
		\psfrag{Y}[r][r][0.8]{$10^9$}
		\psfrag{Z}[r][r][0.8]{$10^{10}$}
		\psfrag{U}[r][r][0.8]{$10^{11}$}
		\psfrag{S}[r][r][0.8]{$10^{12}$}
		\psfrag{T}[r][r][0.8]{$10^{13}$}
		\psfrag{A}[r][r][0.63]{\em SIVF}
		\psfrag{B}[r][r][0.63]{{\em Ding}$^+$}
		\psfrag{C}[r][r][0.63]{{\em Drake}$^+$}
		\psfrag{D}[r][r][0.63]{\em Lloyd-ICP}
		\includegraphics[width=40mm]{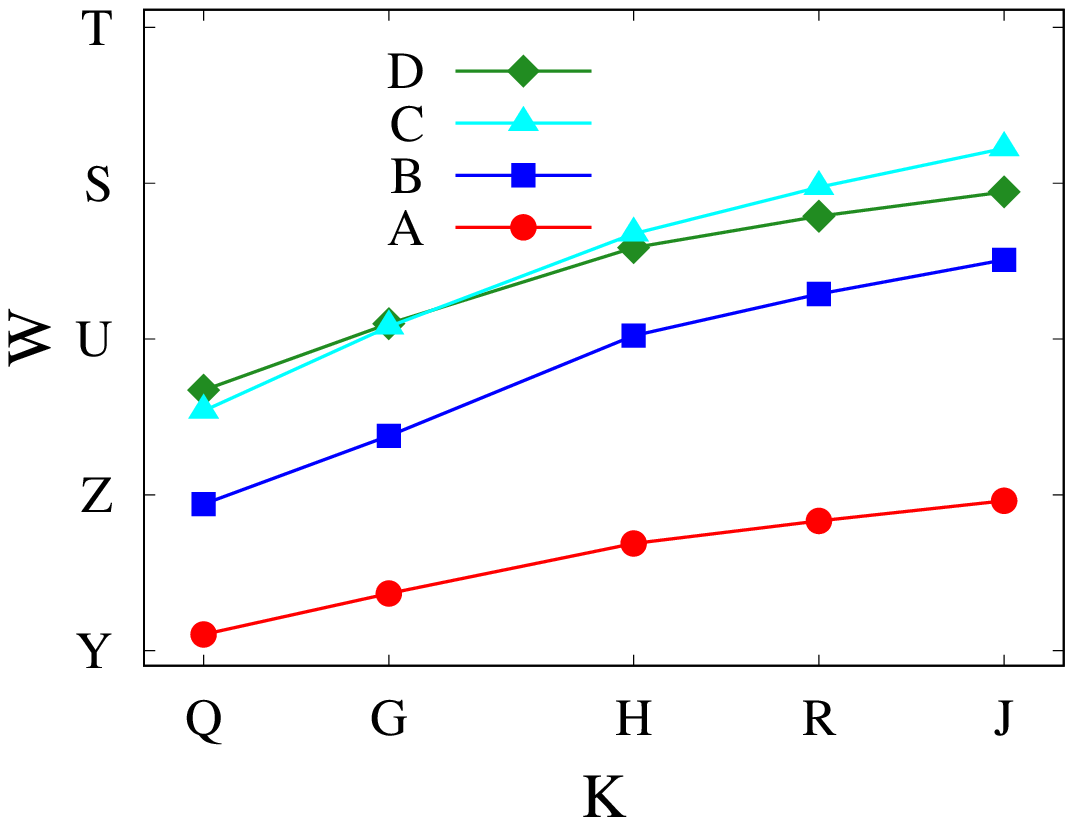}
	}%
\end{center}
\vspace*{-3mm}
\caption{
(a) Number of similarity calculations normalized by (N x k) and 
characteristics of performance degradation factors: (b) the number of instructions (Inst), 
(c) branch mispredictions (BM), and (d) last-level cache misses (LLCM) 
when the three algorithms were executed by 50-thread processing for NYT.
(a) is plotted with log-linear scale and (b), (c), and (d) with log-log scale.
}
\label{fig:comp_dfs_nyt}
\vspace*{-2mm}
\end{figure}

\end{document}